\pdfoutput=1
\documentclass{article}

    \usepackage[preprint]{neurips_2025}
\usepackage{algorithm}
\usepackage{algpseudocode}

\usepackage[utf8]{inputenc} 
\usepackage[T1]{fontenc}    
\usepackage{hyperref}       
\usepackage{url}            
\usepackage{booktabs}       
\usepackage{amsfonts}       
\usepackage{nicefrac}       
\usepackage{microtype}      
\usepackage{xcolor}         
\usepackage{subfig}   
\usepackage{bbding}
\usepackage[font=small,labelfont=bf,tableposition=top]{caption}
\usepackage{float}    
\usepackage{amsmath}
\usepackage{graphicx}
\usepackage{hyperref}  
\usepackage{cleveref}
\usepackage{ntheorem}
\newtheorem{theorem}{Theorem}

\newtheorem{corollary}{Corollary}

\theoremstyle{nonumberplain}
\newtheorem{proof}{Proof}
\title{SceneLCM: End-to-End Layout-Guided Interactive Indoor Scene Generation with Latent Consistency Model}

\author{
  Yangkai Lin\\
  School of Electronic and Information Engineering\\
  South China University of Technology\\
  \texttt{202210182091@mail.scut.edu.cn} \\
  \And
  Jiabao Lei \\
  School of Data Science \\
  The Chinese University of Hong Kong,\\ Shenzhen \\
  \texttt{jiabaolei@link.cuhk.edu.cn} \\
    \AND
    Kui Jia\thanks{Corresponding author.} \\
    School of Data Science \\
    The Chinese University of Hong Kong,\\ Shenzhen \\
    \texttt{kuijia@cuhk.edu.cn} \\
}

\begin{document}

\maketitle

\vspace{-0.6cm}
  \begin{center}
\includegraphics[width=1.0\linewidth]{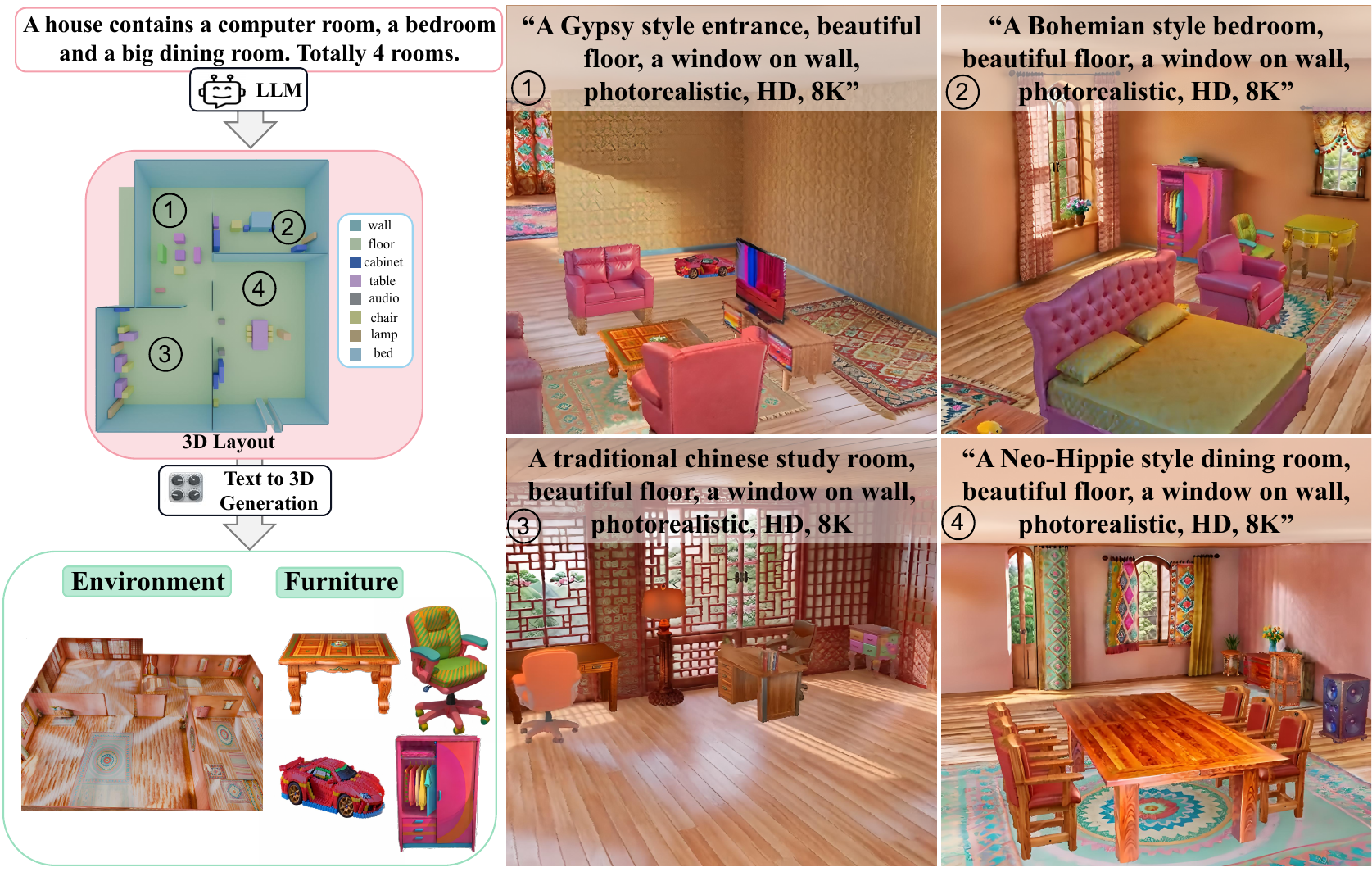}
    \captionof{figure}{Given a textual description of the house, our end-to-end framework enables automated generation of multi-room, multi-scale indoor scene, while supporting controllable scene generation, physically editing, and texture editing of environment. }\label{fig:teaserfigure}
  \end{center}
\vspace{-0.2cm}
\begin{abstract}
\vspace{-0.2cm}
 Our project page: \url{https://scutyklin.github.io/SceneLCM/}. Automated generation of complex, interactive indoor scenes tailored to user prompt remains a formidable challenge. While existing methods achieve indoor scene synthesis, they struggle with rigid editing constraints, physical incoherence, excessive human effort, single-room limitations, and suboptimal material quality. To address these limitations, we propose SceneLCM, an end-to-end framework that synergizes Large Language Model (LLM) for layout design with Latent Consistency Model(LCM) for scene optimization. Our approach decomposes scene generation into four modular pipelines: (1) Layout Generation. We employ LLM-guided 3D spatial reasoning to convert textual descriptions into parametric blueprints(3D layout). And an iterative programmatic validation mechanism iteratively refines layout parameters through LLM-mediated dialogue loops; (2) Furniture Generation. SceneLCM employs Consistency Trajectory Sampling(CTS), a consistency distillation sampling loss guided by LCM, to form fast, semantically rich, and high-quality representations. We also offer two theoretical justification to demonstrate that our CTS loss is equivalent to consistency loss and its distillation error is bounded by the truncation error of the Euler solver; (3) Environment Optimization. We use a multiresolution texture field to encode the appearance of the scene, and optimize via CTS loss. To maintain cross-geometric texture coherence, we introduce a 
normal-aware cross-attention decoder to predict RGB by cross-attending to the anchors locations in geometrically heterogeneous instance. (4)Physically Editing. SceneLCM supports physically editing by integrating physical simulation, achieved persistent physical realism. Extensive experiments validate SceneLCM's superiority over state-of-the-art techniques,
showing its wide-ranging potential for diverse applications.
\end{abstract}
\vspace{-0.8cm}
\section{Introduction}
\vspace{-0.2cm}
While generating diverse indoor scenes is critical for Embodied AI and AR/VR, creating multi-room environments with persistent physical realism remains challenging due to high knowledge requirements and computational costs\cite{blender,unity,unrealengine}. Recent studies have attempted to tackle this problem by developing generative models for scene creation via various approaches, including text-to-3d methods\cite{SceneCraft, Luciddreamer,DreamLCM,Dreamscene,CCD} and Layout-based Object retrieval\cite{holodeck, Anyhome,Architect, InstructScene}. 

Despite these recent efforts, generating diverse, realistic, interactive multi-room environments due to the inherent drawbacks and assumptions made in the pipeline. Contemporary text-to-3D generation frameworks, while advancing high-fidelity asset generation with photorealistic texture details, face systemic challenges in four critical dimensions: (1) Fail to generate multi-room evironments; (2) Inability to decouple furniture components, restricting parametric editing and physical plausibility;(3) Long completion time in large-scale environment; (4) Lack of end-to-end pipeline automation, requiring 3D artists to manually define layouts and camera trajectory\cite{SceneCraft,Dreamscene}. In contrast, Layout-based Object retrieval demonstrate superior scalability and flexible in synthesizing large-scale scene; they also presents its own limitations, (1) Single-step inference paradigms of LLM\cite{gpt3, gpt4, deepseek_r1} often lead to position overlap and orientation inaccuracies; (2) Failing generate texture-rich environments while lacking critical compositional elements (window, ceiling, wall). (3) Necessity of extensive database preparation for inference.

To this end, we propose SceneLCM, an End-to-End generative framework that synergistically integrates text-to-3D and Layout-based generation methods, systematically addressing all aforementioned challenges. We first formulate indoor scene generation pipeline as four modular subtasks: layout generation, furniture generation, environment generation, and physical edition. Specifically, given a text description, we translate it into 3D layout following previous work\cite{Anyhome} as shown in Figure \ref{fig:teaserfigure}. To mitigate void regions and misoriented placement, we propose an iterative programmatic verification mechanism that converts layout parameters into executable programs, followed by LLM-driven iterative refinement cycles where the program undergoes continuous modification and execution until an error-free condition is achieved. Next, we propose Consistency Trajectory Sampling loss(CTS) for furniture generation, which leverage latent consistency model\cite{lcm} to maintain the consistency condition along the PF-ODE trajectory. In detail, the CTS loss is primarily structured around conducting matching between two interval steps in the ODE trajectory. Furthermore, we observe that substituting noise image with rendered image significantly enhances training efficiency while preserving fine-grained texture details. Furthermore, we provide an in-depth theoretical analysis and reveal that the CTS loss and consistency loss\cite{cm} exhibit an explicit mathematical relationship.  

Following the acquisition of finalized furniture assets, we place them into the 3D layout and subsequently optimize the texture of the environment. Prior approaches iteratively inpaint the texture map frame by frame\cite{roompainter, roomtex}, however, this approach results in texture synthesis strictly governed by underlying geometry, and our 3D layout comprising discrete planar meshes invalidates texture inpaint methods. At its core, following SenceTex\cite{scenetex}, we employ a multi-resolution texture field to implicitly represent the texture of the layout. To secure the style consistency and enhance the texture details, we incorporate normal-aware cross-attention decoder and decode RGB values via cross-attending to the pre-sampled anchors scattered across each instance which belongs to the same normal. We further apply CTS loss for texture optimization. Moreover, to enable adaptive multi-scale environment optimization, we propose a zigzag adaptive camera trajectory and several techniques that ensure complete and high-quality texture generation.

We present the first controllable end-to-end multi-room indoor scene generation framework with physically plausible editing support. We show that SceneLCM has the capacity to generate complex scenes that support fully interactable and physical plausible editing. Experimental results demonstrate that our approach generates objects and environments with more details and higher fidelity than SOTA.  


\begin{figure}
    \centering
    \includegraphics[width=1.0\linewidth]{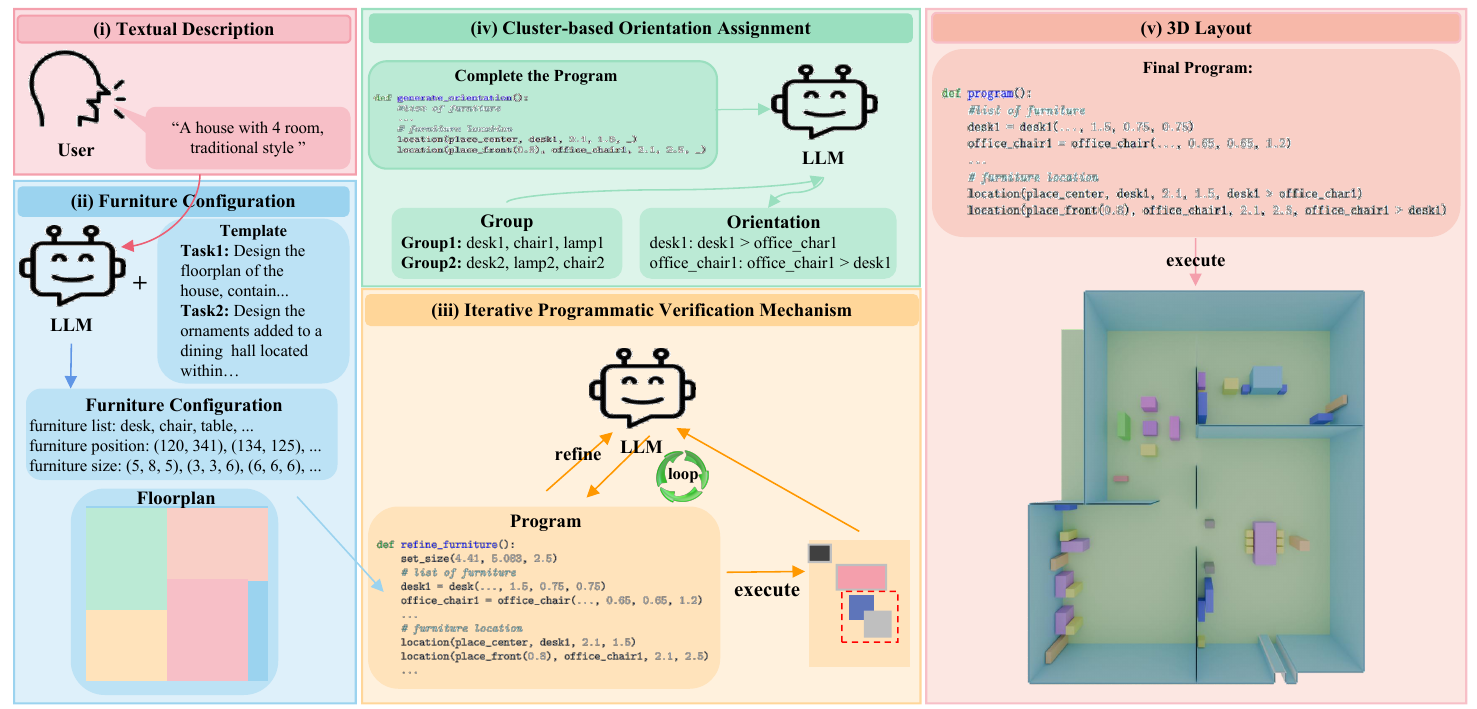}
    \caption{\textbf{The overview of Layout Generation.} Taking a free-form textual input, our pipeline generates the 3Dlayout by: (i)(ii)comprehending and elaborating on the textual prompt through querying an LLM with templated prompts; (iii)transfer the floorplan and furniture configuration into program and apply Iterative Programmatic Verification Mechanism refine the program to achieve the overlap error-free condition; (iv) Organize furniture into groups and generate direction for each group through completing the program; (v)convert the final program into 3D layout.}
    \vspace{-0.3cm}
    \label{fig:llmfigure}
\end{figure}

\vspace{-0.3cm}
\section{Related Works}
\vspace{-0.2cm}
\paragraph{Differentiable 3D Representation}
The differentiable nature of neural rendering paradigms such as NeRF\cite{nerf}, SDF\cite{deepsdf}, and Gaussian Splatting\cite{gaussian_splatting} establishes a unified framework for 3D representation learning, geometric manipulation, and photorealistic rendering. This intrinsic compatibility with gradient-based optimization algorithms enables continuous refinement of volumetric scene parameters through backpropagation, establishing a closed-loop pipeline where loss-driven parameter updates progressively enhance scene fidelity. Particularly, the recent approach\cite{gaussian_splatting}
and its variants\cite{2dgs,mip_splatting,scaffold_gs,sugar} modeling 3D scenes with 3D Gaussians have achieved superior real-time rendering via raseterizer. Unlike implicit representations\cite{nerf, deepsdf}, 3D Gaussians provide more flexible framework, simplifying the integration of multiple scenes. Therefore, we adopt 3D Gaussians for their explicit representation and ease of scene combination.

\paragraph{Text-to-3D Generation}
One work can be categorized as Large Generative Model(LGM)\cite{xcube,lion, direct3d, shape2vecset, clay, triller}. Early LGM leverage generative models\cite{ddpm, gan} for various representations\cite{dpcd,Neural_wavelet, diffusion_nerf}. To enhance both quality and efficiency, recent studies\cite{clay, triller, michelangelo} have resorted to generation in a more compact latent space. However, these methods often yield lacking of texture detail(e.g., hair strands) and unrealistic appearance.

Another line of work can be categorized as text-to-3D generation\cite{CCD,dreamfusion,consistent3d,CFD,Dreamscene, DreamLCM, Luciddreamer, vividdreamer}. As a pioneer, the score distillation sampling (SDS) paradigm for distilling 2D text-to-image diffusion models is proposed in DreamFusion\cite{dreamfusion} and SJC\cite{SJC}. During the distillation process, the learnable 3D representation with rendering is optimized by the gradient to make the rendered view match the given text. Many recent works follow the SDS paradigm and studied for various aspects\cite{dreamtime, hifa, magic3d, prolificdreamer, fantasia3d, Luciddreamer}. Some works\cite{consistent3d, CCD, DreamLCM, vividdreamer} integrating consistency models\cite{cm} exhibit substantial relevance to ours. However, existing approaches merely derive conceptual inspiration from consistency model\cite{cm}, achieving structural analogy without explicitly formalizing the mathematical linkage between consistency function and their methods\cite{consistent3d, CCD, DreamLCM, vividdreamer}. In contrast, we conduct an in-depth analysis of consistency model and mathematically formulate the intrinsic relationship between our CTS loss and consistency loss. 

\paragraph{Indoor Scene Generation}
The Indoor Scene Generation tasks can be categorized into Layout-based Objects retrieval and text-to-3D Generation. The first methods mainly focus on layout generation and then retrieve objects from database according to layout. In the early stage, some works\cite{InstructScene, echoscene} train generative models\cite{gan, ddpm} to generate layout. However, the 3D scene datasets are considerable small compared with object datasets. Thus models trained on these datasets are less robust and lack of novelty. Recently, some works\cite{Anyhome, holodeck, globalTree} utilize the LLM/VLM\cite{gpt3, gpt4} to analysis user text prompts and use CLIP to retrieve relevant objects. Though these methods generate editable scenes, they often neglect texture optimization, require annotated database, and omit key scene elements (e.g., window, ceiling, wall). Text-to-3D scene generation methods face significant limitations. \cite{text2nerf, text2room, controlroom3d} replying on inpainted images for scene completion can generate realistic visuals but suffer from limited 3D consistency. Moreover, such methods typically do not allow for the editing of furniture. While some methods\cite{Dreamscene, SceneCraft}, akin to ours, attempt to merge layout to enhance controllability, and sds\cite{dreamfusion} loss for realistic scene. However, they are incapable of autonomously generating layouts and can accommodate only a limited number of objects. Moreover, They cannot perform physically plausible editing. Our method addresses all the aforementioned issues and proposes an end-to-end, controllable, high-detail  generative framework that supports physically plausible editing.
\begin{figure}
    \centering
    \includegraphics[width=1.0\linewidth]{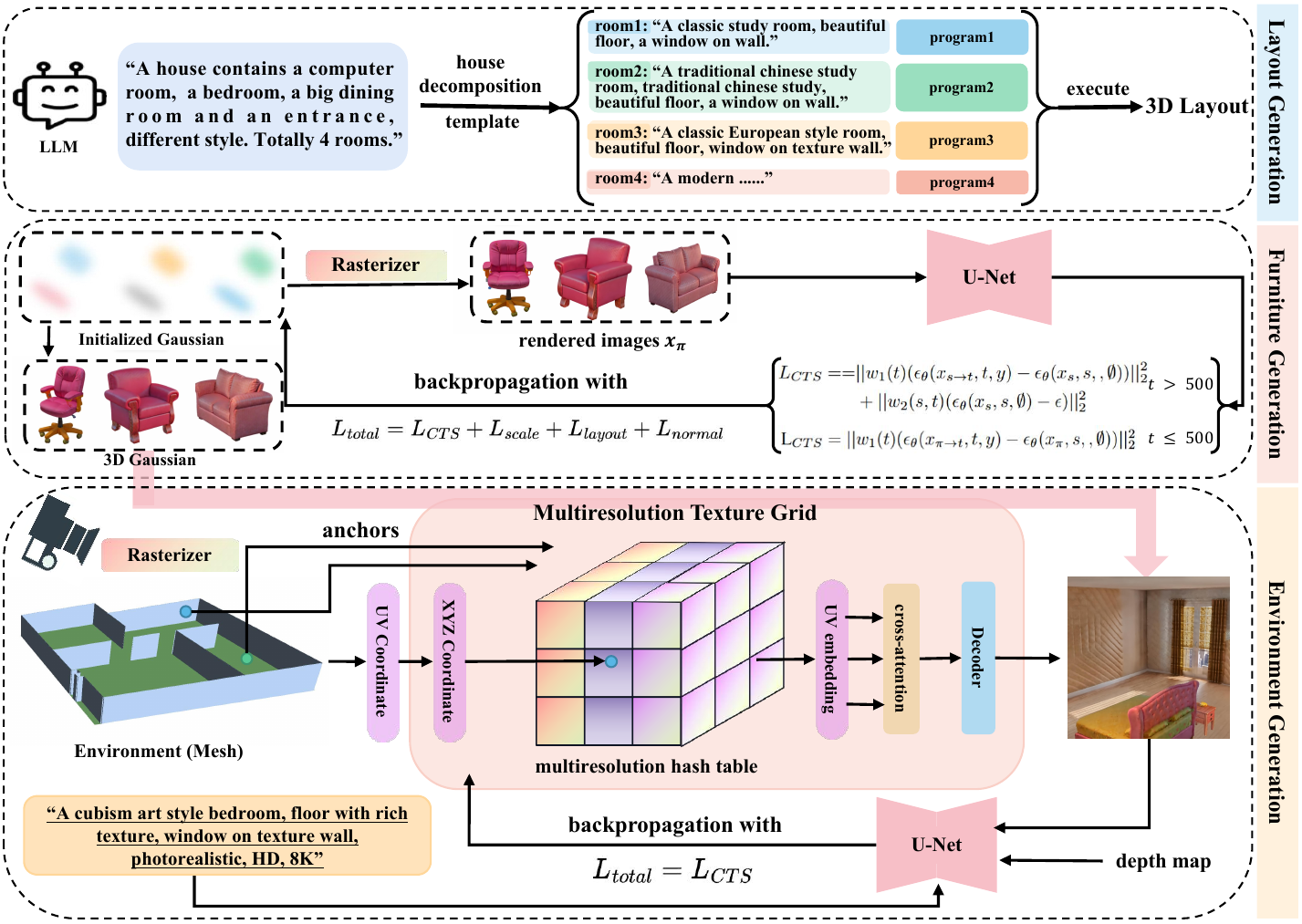}
    \caption{\textbf{The overview of SceneLCM.} We primarily employ CTS loss to jointly generate high-quality furniture and optimize environment appearance. Additionally, SceneLCM ensures scene-wide consistency through camera sampling and allows for flexible editing by integrating furniture with the environments in the scene.}
    \label{fig:pipeline}
    \vspace{-0.3cm}
\end{figure}

\vspace{-0.3cm}
\section{Preliminary}
\vspace{-0.2cm}
\paragraph{Consistency Model(CM)\cite{cm, lcm}} is proposed to facilitate a single-step or low number of function evaluations (NFEs) generation by distilling knowledge from pre-trained DM\cite{ddpm}. It defines a consistent function $f_\theta(.;.)$ with trainable parameters $\theta$ that directly predicts the denoised image $x_0$ given $t$ and $x_t$. And $f_\theta(.;.)$ is trained by minimizing self-consistency distillation loss defined as:
\begin{equation}\label{CM_loss}
    L_{CD}(\theta, \theta^-) = \mathbb{E} [w(t)||f_{\theta}(x_{t_{n+1};t_{n+1}}) - f_{\theta^-}(\hat{x}_{t_{n+1} \rightarrow t_n};t_n)||_2^2],
\end{equation}
where $0 = t_1 < t_2 ... < t_N = T$, $\hat{x}_{t_{n+1} \rightarrow t_n};t_n)$ is calculated given ODE solver $\Phi(.)$. $\theta^-$ is updated during training process through an exponential moving average(EMA) strategy. The ultimate goal of CM is to maintain the self-consistency condition along the trajectory$\{x_t\}_{}{t \in [0, T]}$, satisfying: 
\begin{equation}\label{CM_condition}
    f(x_t; t) = f(x_{t'};t') \quad \forall t, t' \in [0, T],
\end{equation}

Specifically, consistent function $f$ parameterized by the noise predition model $\epsilon_\theta$, as follows:
\begin{equation} \label{CM_function}
    f(x_t, t) = c_{skip}(t) x_t + c_{out}(t) (\frac{x_t - \sigma_t \epsilon_\theta(x_t, t)}{\alpha_t}),
\end{equation}

Inspired by CM\cite{cm}, we propose Consistent Trajectory Sampling to guide the rendered images $x_\pi$ to match the origin of PF-ODE trajectory.

\vspace{-0.2cm}
\paragraph{3D Layout}
3D layout acts as a rough outline for the house layout and is defined by semantic bounding boxes which represent layout geometry using mesh. Specifically, we define a house $M_e$ as a set of semantic bounding boxes $B = (p, s, c, n)$, where $p \in \mathbb{R}^3$ is the box center, $s \in \mathbb{R}^3$ is the box size, $c$ the semantic class id, and $n$ the unique name associated with each box. As shown in figure. \ref{fig:llmfigure}, 3D layout is solely composed of planar components with limited geometric complexity. 

\vspace{-0.2cm}
\paragraph{Hybrid Rendering} 


Scene layouts combine low-complexity geometry with high-frequency texture and we represent layout geometry using mesh and texture through multiresolution texture field, while parameterizing furniture via 3D Gaussians. We employ a differential rasterizer (i.e., pytorch3d\cite{pytorch3d}) to render rgb image $I_{l}$ and depth map $D_{l}$ from layout mesh, and Gaussian splatting render\cite{gaussian_splatting} to render rgb image $I_{f}$ and depth map $D_f$ from furniture. The renderings of the scene are obtained by fusing the renderings of furniture and layout, as follows:
\begin{equation}
R = \begin{cases}
     R_{f} & \text{if } D_{f} \leq D_{l} \\
    R_{l}  & \text{if } D_{f} \ge D_{l}
\end{cases}
\end{equation}
where $R$ denotes the final RGB image $I$, $R_l$ is the pixel value from $I_l$ and $R_f$ is from $I_f$ respectively. 

\vspace{-0.3cm}
\section{Method}
\vspace{-0.2cm}
SceneLCM formulates the indoor scene generation pipeline as four modules: Layout Generation, Object Generation, Environment Optimization, and Physical Editing. First, we modulate the textual input using LLM\cite{gpt4} as detailed in Sec. \ref{layout_generation}, which involves user input comprehension and some methodologies for fine-grained layout parameter refinement. Second, in Sec. \ref{furniture_generation}, we rapidly create high-fidelity furniture using CTS loss by incorporating LCM\cite{lcm}, accompanied by two theoretical justification. Third, Sec. \ref{environment_optimization} describes texture field and optimization. Finally, we show that our generated indoor scene can support physical plausible editing, texture editing in Sec. \ref{Qualitative}. 
 \vspace{-0.2cm}
\subsection{Layout Generation} \label{layout_generation}
\vspace{-0.2cm}

\begin{figure}
    \begin{minipage}[t]{0.5\textwidth}
    \vspace{0pt}
    \centering
\includegraphics[width=1\linewidth]{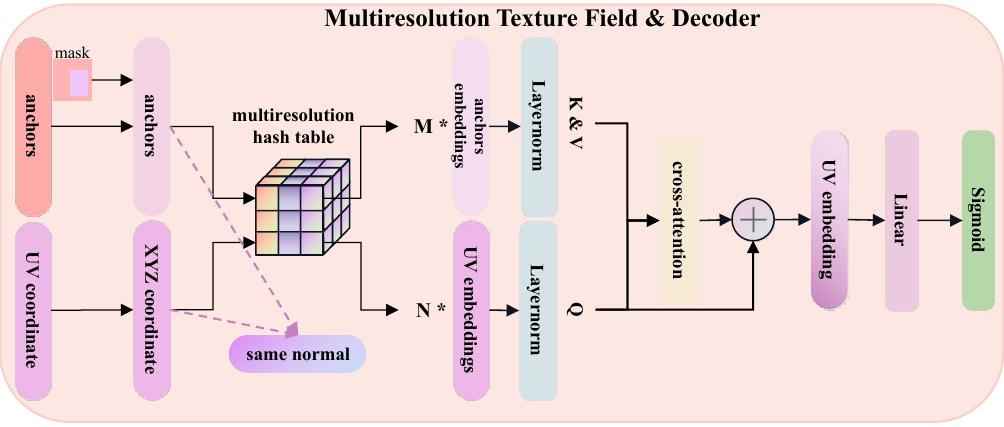}
        \caption{\textbf{Normal-aware Texture Decoder.} We mask out the anchors which have the same normal as the UV coordinate and extract the texture embedding for the XYZ and anchors. Then we employ a cross-attention to produce the final UV embeddings.}
  \label{fig:decoder}
    \end{minipage}
    \hfill
    \begin{minipage}[t]{0.48\textwidth}
    \vspace{1cm}
  \captionof{table}{Quantitative Results of SceneLCM compared with baselines}
  \label{quantitive_comparision}
  \centering
  \resizebox{\textwidth}{!}{
    \begin{tabular}{@{}c@{\hskip 0.5em}c@{\hskip 0.5em}c@{\hskip 0.5em}c@{\hskip 0.5em}c@{\hskip 0.5em}c@{}}
    \toprule
    Methods   & Editing     & Physics & Multi-room & Time (h) & User Study\\
    \midrule
    Text2Room~\cite{text2room} &   &  &  &  9.1 & 4.1   \\ 
    Set-the-Scene~\cite{set_the_scene}     & \Checkmark &  &   & 1.6 & 3.2    \\
    DreamScene~\cite{Dreamscene}     & \Checkmark       &  &  & \textbf{1.2} & \textbf{7}  \\
    SceneCraft~\cite{SceneCraft} &   &  & \Checkmark &  4.5 & 4.4 \\
    Ours & \Checkmark  & \Checkmark & \Checkmark & \textbf{1.2} & \textbf{8.4} \\
    \bottomrule
  \end{tabular}
  }
    \end{minipage}
    \vspace{-0.5cm}
\end{figure}
This initial component of our framework performs four key functions as shown in Figure \ref{fig:llmfigure}. First, user offer a description of the house to guide and control the generation. Second, it utilizes the LLMs\cite{gpt3, gpt4, deepseek_r1, deepseek_v3} to convert the provided input into a 3D layout encompassing house floorplan
and furniture configurations\cite{Anyhome}. However, the layouts generated by this simplistic approach suffer from blank area, position overlap and orientation inaccuracies. Third, we propose an iterative programmatic verification mechanism that convert the 3D layout into programs and execute it to check for conflicts. Then we iteratively feed the error and program into the LLM for parameter refinement, cycling this process until error-free conditions are achieved.  Moreover, we propose a cluster-based orientation assignment strategy, where scene objects are partitioned into functional groups. Within each group, object 
  orientations are determined not by cardinal directions (e.g., north/south), but through inter-object spatial relations (e.g., chair $\rightarrow$ desk). LLMs can efficiently infer object orientations with minimal local context. Further details and algorithm on these prompts can be found in the Supplementary Material.
\vspace{-0.2cm}
\subsection{Furniture Generation} 
\vspace{-0.1cm}
\label{furniture_generation}

\paragraph{Consistency Trajectory Sampling Loss}.

Consistent3D\cite{consistent3d} argues that the key to generating a satisfactory 3D model is to accurately perform the 3D ODE sampling using diffusion model. Motivated by this assumption, recent works\cite{CCD, DreamLCM, vividdreamer} attempt to improve the model's performance in view of CM\cite{cm}. However, these models only adopt the idea of consistency model, but fail to conduct in-depth research on their interrelationships. 

Similar to eq. \ref{CM_loss}, our Consistency Trajectory Loss is derived as follows:
\begin{equation}
    \begin{aligned}
        L_{CTS} &= \mathbb{E}[||w_1(t)(\epsilon_\theta(x_{s \rightarrow t}, t, y) - \epsilon_\theta(x_{s}, s, , \emptyset))||_2^2 + 
    ||w_2(s, t)(\epsilon_\theta(x_{s}, s, \emptyset) - \epsilon)||_2^2] \\
    w_1(t) &= c_{out}(t)(\frac{\sigma_{t}}{\alpha_{t}}) \\
    w_2(s, t) &= [c_{out}(t) - c_{out}(s)](\frac{\sigma_{s}}{\alpha_{s}})
    \end{aligned}
\end{equation}
where $t > s$ are two adjacent diffusion time steps, $x_s = \alpha_s x_\pi + \sigma_s \epsilon$, and $x_{s \rightarrow t}$ is a less noisy sample derived from deterministic sampling by running one discretization step of a ODE solver from $x_s$. $\epsilon$ is the random noise that will only be sampled once and kept fixed. $y$ is the text condition.

\paragraph{Justification.} In the following, we offer two theoretical justifications to demonstrate that, our CTS loss is equivalent to the Consistency Loss\cite{cm} and upon achieving convergence, our CTS loss is capable of generating a high-fidelity 3D model.

\begin{theorem} \label{the1}
    Let $f_\theta(\cdot)(x, t)$ denote the pre-trained consistency function. We assume $f_\theta(\cdot)$ satisfies the formulation defined in Latent Consistency Model\cite{lcm}, and $t \ge 30$. Assume further that for all $t \ge 30$, the ODE solver $G$ called at $t_{n+1}$ has local error uniformly bounded by $O((\triangle t)^{p+1})$ with $p \ge 1$,  The Consistency Loss\cite{lcm}  can be mathematically expressed as the sum of the Consistency Trajectory Sampling Loss and one infinitesimal components, along with a term whose magnitude is bounded by $10^{-7}$:
    \begin{equation}
        \begin{aligned}
            L_{CD} &= L_{CTS} + \underbrace{(-\frac{O((\triangle t)^2)}{\alpha_{t_{n+1}}}) + c_{skip}(t_{n+1})O((\triangle t)^{p+1})}_{\text{infinitesimal components}} + \underbrace{m}_{|m| \le 10^{-7}} \\
            &= L_{CTS} + O((\triangle t)^2) + m
        \end{aligned}
    \end{equation}
\end{theorem}
where $c_{skip}(\cdot), c_{out}(\cdot)$ is coefficient in consistency function.

\begin{theorem} \label{the2}
Assume that the pre-trained noise predictor $\epsilon_\theta(\cdot;\cdot)$ in Consistency Model\cite{lcm, cm} satisfies the Lipschitz condition. Define $\triangle := sup\{|t_{n} - t_{n+1}|\}$. For any given camera pose $\pi$, if convergence is achieved according to $L_{CTS}$, then there exists a corresponding real image $x_0 \sim p_{data}(x)$ such that:
\begin{equation}
    ||x_\pi - x_0||_2 = O(\triangle)
\end{equation}
where $x_\pi = g(\theta, \pi)$ denotes the rendered image for pose $\pi$.
\end{theorem}
In addition, we can proof that the existing methods\cite{vividdreamer, DreamLCM} which incorporate Consistency Model\cite{cm} can be amalgamated into our framework. We provide the full proof in Appendix. 

Theorem \ref{the1} guarantees that the CTS loss function comprehensively inherits the capability of the Consistency model, while Theorem \ref{the2} demonstrates that images synthesized by our framework exhibit both photorealism and semantic alignment with real-world environments.

Once 3DGS developed distinct semantic signal which align with the text prompts, adding noise drives the model to alter details for text alignment. However, as most text lack high-frequency specifics, the focus should be on enhancing detail generation rather than forces alignment. Therefore, we remove the noise and the second term of $L_{CTS}$, and set $x_s = x_\pi$ when $t \le 500$ as shown in Figure \ref{fig:pipeline}. To constrain the furniture to maintain scale, position align with the provided layout priors, we introduce the bounding box loss:

\begin{equation}
    \begin{aligned}
        L_{Layout} =&{} d^x(G_x, x_i, h_i) + d^y(G_y, y_i, w_i) + d^z(G_z, z_i, l_i) \\
        d^x(G_x, x_i, h_i) =&{} ||\min{(G_x)} - (x_i - \frac{h_i}{2})||_2^2 + || \max{(G_x)} - (x_i + \frac{h_i}{2})||_2^2    
    \end{aligned}
\end{equation}
where $G_x$ is a set of furniture Gaussian center $\mu$ on the x-axis, $x_i$ is the position of the layout center on the x-axis, and $h_i$ is the height of the layout prior to the i-the furniture. Additionally, we incorporate $L_{scale}$ and $L_{normal}$ for texture detail enhancement. Further details be found in the Supplementary Material.
\vspace{-0.2cm}
\subsection{Environment Optimization} 
\vspace{-0.2cm}

\label{environment_optimization}
After obtaining furniture, we place the furniture into 3D layout and start optimizing environment. Previous methods directly optimize latent map with the SDS loss\cite{scenetex} or iteratively inpaint the missing texture in each viewpoint. While optimization-based approaches often suffer from multi-view geometric inconsistencies due to gradient conflicts across viewpoints, inpainting-based methods exhibit limitations in synthesizing high-frequency texture detail and geometrically plausible ornaments (e.g., wall moldings, cornices). In this section, we begin by introducing a multi-resolution texture field module and a normal-aware decoder to synthesize visually coherent and spatially consistent texture. Subsequently, we propose a zigzag adaptive camera trajectory to address environment optimization across multiple spatial scales and apply CTS loss for realistic texture generation.
\begin{figure}
    \centering
\includegraphics[width=1.0\linewidth]{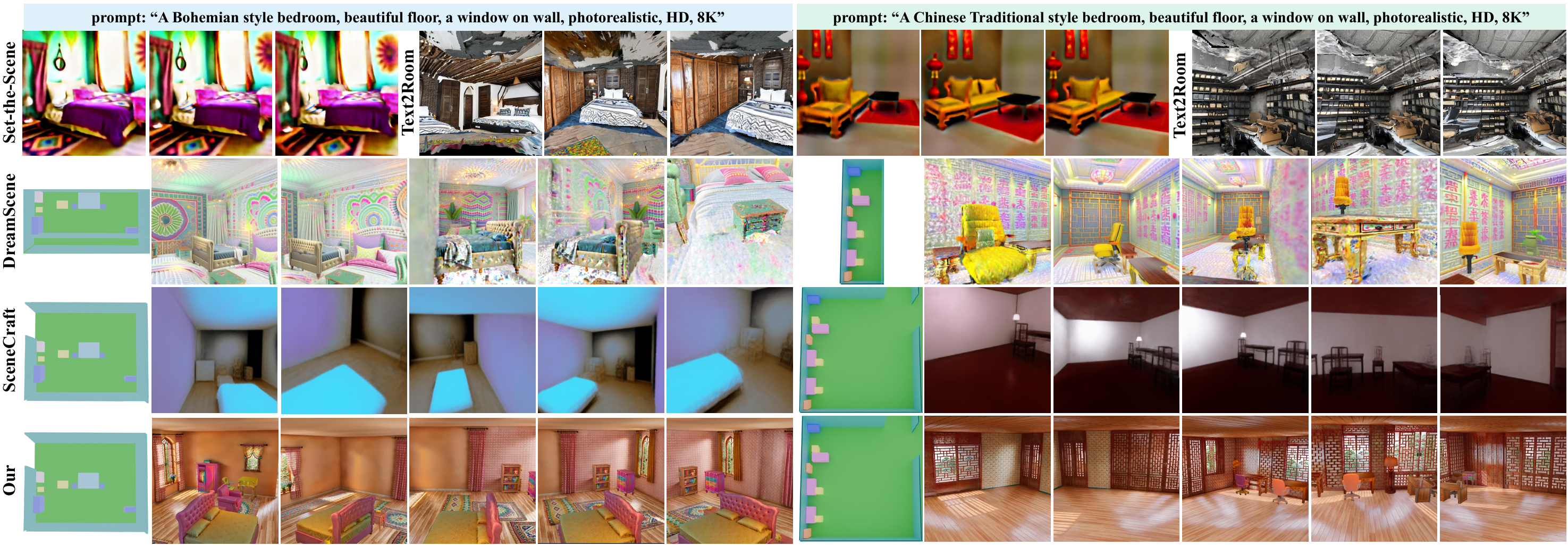}
    \caption{Qualitative comparisons of SceneLCM and baselines. Set-the-Scene\cite{set_the_scene} and Text2room\cite{text2room} generate incomplete results; 
    The floorplan of DreamScene\cite{Dreamscene} is automatically generated and the enviroment exhibits multi-view inconsistency, with floor and bed fused together in column five; SceneCraft\cite{SceneCraft} generates incorrect style outputs and blurry images. Our method is capable of producing higly detailed scenes, including realistic floor textures, while simultaneously generating wall decorations such as curtains and windows.}
    \label{fig:comparison_all}
    \vspace{-0.5cm}
\end{figure}

\vspace{-0.2cm}
\subsubsection{Texture Field \& Decoder}
\vspace{-0.1cm}
To tackle those inherent disadvantages, we adopt a multi-resolution texture field that queries the texture features with given UV coordinates\cite{scenetex}. We encode texture features for all UV coordinate $q$ 
    at each scale, and concatenate those features as the output UV embeddings $ \varepsilon(q)$ to faithfully represent all texture details. Then the UV embeddings are decoded to the final RGB value by the cross-attention texture decoder. Motivated by the empirical observation that consistent normal correlate with texture style divergence, inconsistent normal correlate with texture style consistency, we propose an effective rendering module with normal awareness to predict RGB values. Specifically, for each rasterized UV coordinate, we apply a UV instance mask to mask out instance with the same normal. Then, we obtain the UV embeddings for the rasterized locations in the view. Meanwhile, we extract the texture features for the pre-sampled UVs scattered across this instances with same normal as the reference UV embeddings as shown in Figure. \ref{fig:decoder}. We deploy a multi-head cross-attention module to produce the instance-aware UV embeddings. Followed SceneTex\cite{scenetex}, we treat the rendering UV embeddings as Query, and the reference UV embeddings as Key and Value. Finally, a shared MLP projects the UV embeddings to RGB values. We denote the whole rendering process as $C = M(\varepsilon(q);\phi)$, where $C$ represents an RGB image arbitrary resolution, $M(\cdot;\phi)$ is a differentiable function resembles the entire texture field and decoder with trainable parameters $\phi$. 
\begin{figure}
    \centering
    \includegraphics[width=1.0\linewidth]{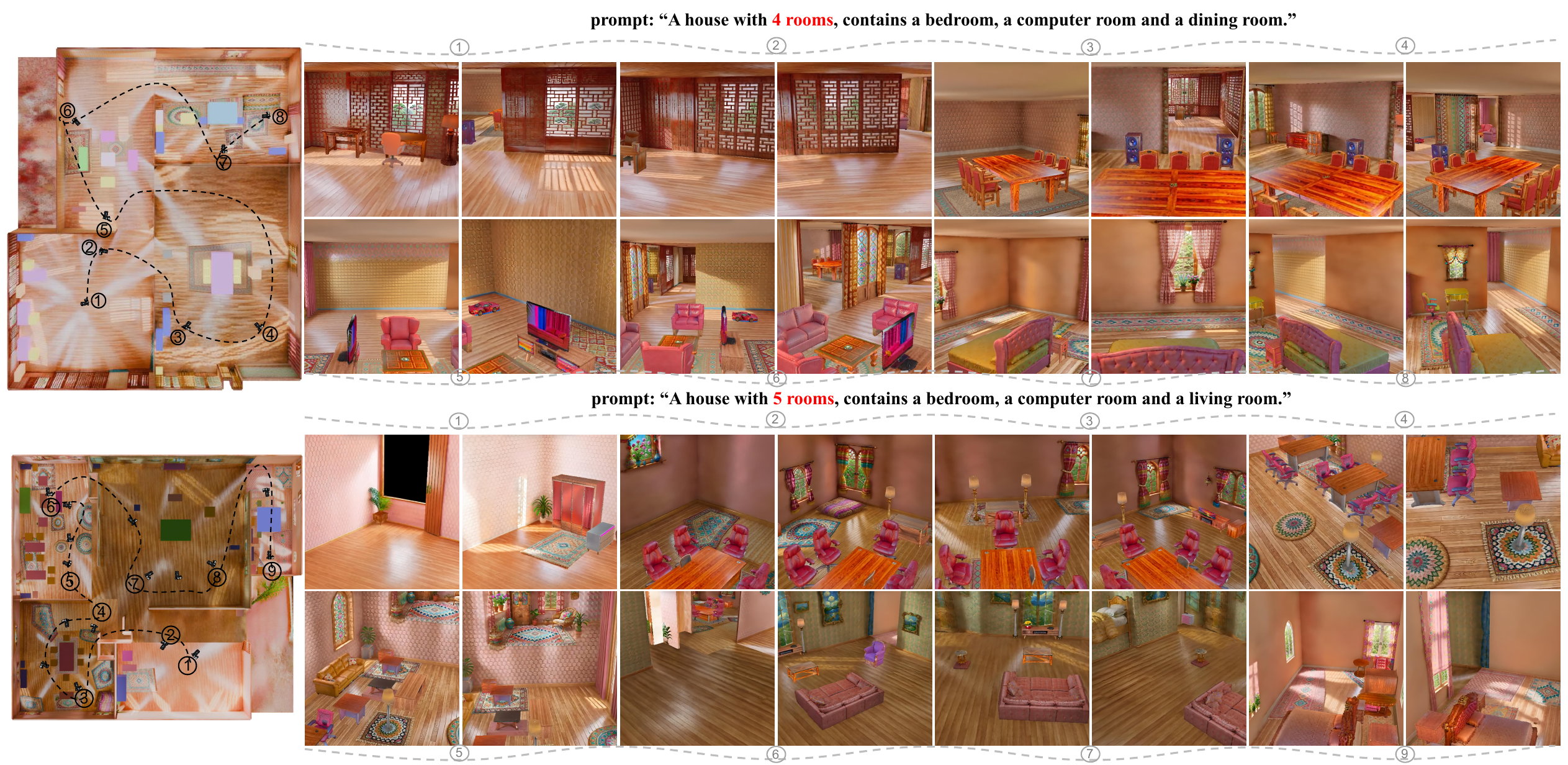}
    \caption{Generation results of SceneLCM in multi-room. The left column represents the layout, where we distill the multi-resolution texture field on a 4096×4096 texture map serving as the texture map of layout. We demonstrate SceneLCM’s ability to generate more complex indoor scenes.}
    \label{fig:full_layout}
    \vspace{-0.3cm}
\end{figure}
\vspace{-0.2cm}
\subsubsection{Texture Optimization}
\vspace{-0.2cm}
We adopt a LCM\cite{lcm} as a critic to optimize the texturing module following the strategy of SceneTex\cite{scenetex}. We render RGB image via querying the texture field $C = M(\varepsilon(q); \phi)$. In each iteration, optimize $C$ via the CTS objective with a pre-trained frozen depth-conditioned diffusion $\epsilon_\theta$:
\begin{equation}
        \nabla_\phi L_{CTS} = \mathbb{E}_{t, \epsilon}[(|w_1(t)(\epsilon_\theta(x_{s \rightarrow t}, t, y, x_\pi^d) - \epsilon_\theta(x_{s}, s, , \emptyset))| + 
    |w_2(s, t)(\epsilon_\theta(x_{s}, s, \emptyset) - \epsilon)| )\frac{\partial M(\varepsilon(q);\phi)}{\partial \phi}]
\end{equation}
where $x_s = \alpha_s x_\pi + \sigma_s \epsilon$, $x_\pi = M(\varepsilon(q);\phi)$ and $x_\pi^d$ is the depth map of $x_\pi$. We also remove the noise after $t \le 500$.

Unlike previous methods\cite{scenetex, roomtex, roompainter}, we focus on multi-room settings. Direct application of CTS loss often introduces patchy artifacts and incomplete texture maps, stemming from ultra-dense UV parameterization causing adjacent UV coordinates to exhibit tight numerical proximity and unstable camera trajectories. These numerical proximities induce pathological initializations, leading to local optima and unstable camera trajectories that fail to capture global scenes at large scales.

 To address those issues, we treat UV coordinates as spherical coordinates and convert them to cartesian coordinates, while applying layernorm after each UV embedding to amplify inter-data distinctions. We propose a zigzag adaptive camera trajectory where the camera and target closely follow walls during motions. Specifically, the camera's xy coordinates move in the opposite direction to those of the target, while its height is inversely proportional to the target's height. 

\vspace{-0.3cm}
\section{Experiments}
\vspace{-0.2cm}
\paragraph{Implementation Details.} We utilized GPT-4\cite{gpt4} as our LLM for 3D layout generation, latent consistency model\cite{lcm} as our diffusion model, Point-E for initial representation of objects. We tested SceneLCM and all the baselines one the A800 GPU for fair comparison.
\vspace{-0.2cm}
\paragraph{Baselines.} For the comparison of the scene generation, we use the current open-sourced SOTA methods Text2Room\cite{text2room}, set-the-scene\cite{set_the_scene}, DreamScene\cite{Dreamscene} and SceneCraft\cite{SceneCraft}. For text-to-3D generation, we select open-source SOTA methods luciddreaner\cite{Luciddreamer}, vividreamer\cite{vividdreamer}, CCD\cite{CCD} and DreamScene\cite{Dreamscene}.

\begin{figure}
    \begin{minipage}[t]{0.5\textwidth}
    \vspace{0pt}
    \centering
\includegraphics[width=1\linewidth]{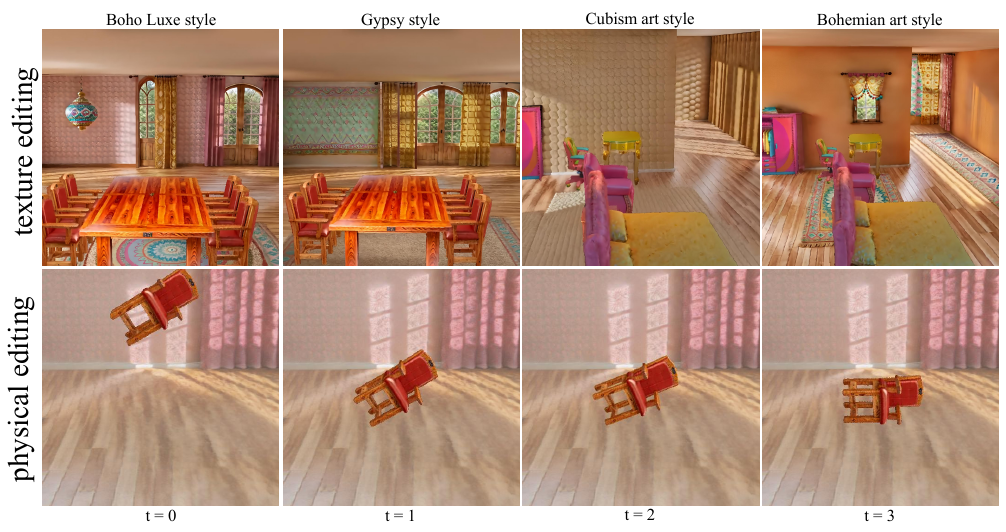}
        \caption{Editing Results.}
  \label{fig:application}
    \end{minipage}
    \vspace{-0.3cm}
    \hfill
    \begin{minipage}[t]{0.48\textwidth}
    \vspace{0pt}
        \centering
\includegraphics[width=1\linewidth]{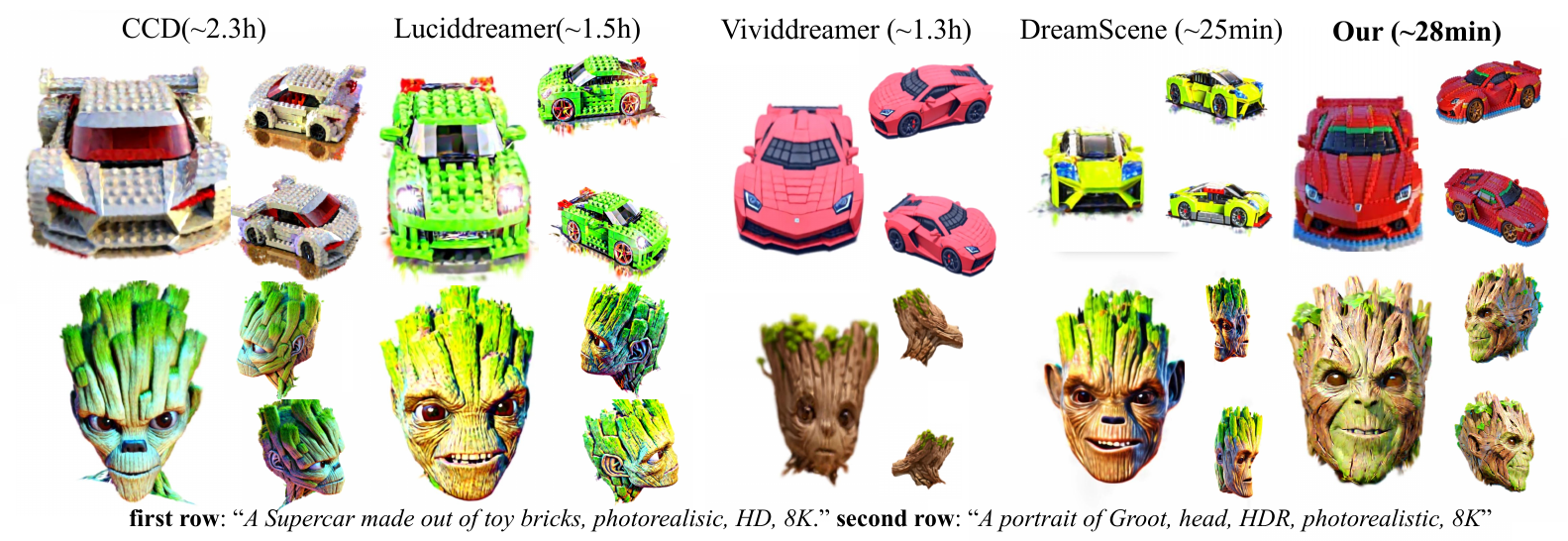}
        \caption{Comparison with baselines in text-to-3D generation.}
  \label{fig:obj}
    \end{minipage}
    \vspace{-0.3cm}
\end{figure}
\vspace{-0.2cm}
\paragraph{Evaluation Metrics}
We tested the generation time of each method\cite{text2room, set_the_scene, Dreamscene, SceneCraft}, compared the editing and generative capabilities against their published papers,
and did a 20-participant user study to score(out of ten) the quality of the videos generated by each method for 3 scenes of 15 seconds.

\vspace{-0.2cm}
\subsection{Qualitative Results}\label{Qualitative}
\vspace{-0.1cm}
\paragraph{Indoor Scene.} Figure \ref{fig:comparison_all} shows the comparison of SceneLCM with the SOTA. We rotated the camera to capture a 360-degree scene, demonstrating our consistency. Text2Room\cite{text2nerf} uses the text-conditioned inpainting model to generate frame by frame and only produces satisfactory results under camera poses encountered. Set-the-Scene\cite{set_the_scene} is a Nerf-composition methods that lacks the capability to generate objects with varying scales, resulting in blurred results. SceneCraft\cite{SceneCraft} finetunes a 2D diffusion model conditioned on bounding-box images, and apply a SDS loss for scene optimization. Our generated layouts are incompatible with SceneCraft's native format, so we create layouts for SceneCraft manually to align with ours as closely as possible. Due to the insufficient capacity of the SceneCraft-finetined diffusion model, it struggles to reason the layouts which contain many closely placed objects. DreamScene\cite{Dreamscene} employs Formation Pattern Sampling for objects generation, and a progressive three-stage camera sampling strategy to effectively ensure object and environment integration. DreamScene exclusively processes relative spatial relationships between objects to autonomously generate floorplans. Due to DreamScene\cite{Dreamscene} lacks of geometric constraints in environment, the resulting environments typically exhibit multi-view inconsistency. For example, in the second row of the sixth column of Figure \ref{fig:comparison_all}, the floor and bed are seamlessly integrated into a unified structure. Additionally, the unrestricted proliferation of Gaussians induces unrealistic visual artifacts. In contrast, our model addresses all these issues: It can generate complex and realistic scenes while ensuring multi-view consistency.

\vspace{-0.2cm}
\paragraph{Objects.} Figure \ref{fig:obj} shows that our CTS can generate realistic 3D representation following the text prompt in a short time. DreamScene\cite{Dreamscene} is fastest yet at the cost of low generation quality.
\paragraph{More Results.} In Figure \ref{fig:full_layout}, we showcase the results of multi-room generation with full layout on the free-camera trajectory. Users only need to input text, and our end-to-end model autonomously generates entire houses with distinct styles per room, requiring zero human intervention throughout the entire generation process.

\vspace{-0.2cm}
\paragraph{Editing.} 
In Figure \ref{fig:application}, we show 4 variants of generation with two room layouts and different appearance, simply achieved by using different texture map of environment in first row. In second row, we show that our model can support physically plausible editing. A chair in mid-air will fall downward under gravitational force.

\vspace{-0.2cm}
\subsection{Quantitative Results}
\vspace{-0.2cm}
Since baselines\cite{SceneCraft, text2room, set_the_scene} are not able to generate objects in the environment independently, for a fair comparison we calculate the generation time of our environment generation stage per room. The left side of Table \ref{quantitive_comparision} shows that we have robust editing capabilities and the right side shows the user study, where SceneLCM is far ahead of baselines in terms of quality.

\vspace{-0.2cm}
\subsection{Ablation Study}
\vspace{-0.2cm}
The appendix section offers a comprehensive introduction to all of our evaluated models and additional
experimental details. Please refer to
Appendix for more details.
\vspace{-0.3cm}
\section{Limitation and Conclusion}
\vspace{-0.2cm}
SceneLCM currently exhibits limitations in generating high-fidelity layout arrangements (e.g., books on shelves, cups on tables) as Architect\cite{Architect}. Secondly, our CTS loss still suffers from the Janus issue, causing dualistic artifacts.

In summary, we introduce an novel end-to-end indoor scene generative framework. By employing layout refine strategy, Consistency Trajectory Sampling(CTS), texture field and normal-aware decoder for environment optimization, we address the current issues of inefficiency, inconsistency, single-room generation, and limited physical editability in current text-to-3D scene generation methods. Extensive experiments have demonstrated that SceneLCM is a milestone achievement in 3D scene generation, holding potential for wide ranging applications across numerous fields.

\bibliographystyle{plainnat}
\bibliography{neurips_2025.bib}

\newpage
\section{Supplement Detail}

\begin{itemize}
    \item Theoretical Proof \ref{Theoretical}
    \item Pipeline Detail \ref{Pipeline}
    \item Discussion for Related Work \ref{Discussion for Related Work}
    \item Explanation of Experiments \ref{Explanation of Experiments}
    \item Additional Experiments \ref{additional_experiments}
    \item Ablation Study \ref{ablation_experiments}
\end{itemize}

\subsection{Theoretical Proof}
We first prove two theorems and one corollary, then conduct ablation studies to demonstrate that the $m$ exerts no influence on the experimental results.

\subsubsection{Theoretical Proof} \label{Theoretical}
\begin{theorem} \label{the1}
    Let $f_\theta(\cdot)(x, t)$ denote the pre-trained consistency function. We assume $f_\theta(\cdot)$ satisfies the formulation defined in Latent Consistency Model\cite{lcm}, and $t \ge 30$. Assume further that for all $t \ge 30$, the ODE solver $G$ called at $t_{n+1}$ has local error uniformly bounded by $O((t_{n+1} - t_{n})^{p+1})$ with $p \ge 1$,  The Consistency Loss\cite{lcm}  can be mathematically expressed as the sum of the Consistency Trajectory Sampling Loss and two infinitesimal components, along with a term whose magnitude is bounded by $10^{-7}$:
    \begin{equation}
        \begin{aligned}
            L_{CD} &= L_{CTS} + \underbrace{(-\frac{O(h^2)}{\alpha_{t_{n+1}}}) + c_{skip}(t_{n+1})O((\triangle t)^{p+1})}_{\text{infinitesimal components}} + \underbrace{m}_{|m| \le 10^{-7}} \\
            &= o((\triangle t)^2) + m \\
            h &= log(\frac{\alpha_{t_n}}{\sigma_{t_n}}) - log(\frac{\alpha_{t_{n+1}}}{\sigma_{t_{n+1}}})
        \end{aligned}
    \end{equation}
\begin{proof}
The proof is based on the formulation defined in Latent Consistency Model\cite{lcm}. We have $f_\theta(x, t) = c_{skip}(t)x + c_{out}(t) F_\theta (x_t, t) $, where $F_\theta (x_t, t) = \frac{x_t - \sigma_t \epsilon_\theta(x_t, t)}{\alpha_t}$, $c_{skip}(t) = \frac{\sigma^2}{(\frac{t}{0.1})^2 + \sigma^2}$, $c_{out}(t) = \frac{\frac{t}{0.1}}{\sqrt{(\frac{t}{0.1})^2 + \sigma^2}}$, and $\sigma = \frac{1}{2}$. $x_t$ is obtained by ODE solver applied to $x_s$. To simplicity, we omit the condition $c$ of the consistency function and exponential moving average(EMA) of the parameter $\theta$. We use the DPM-solver as ODE solver $G(x_s, s, t)$, $x_{s \rightarrow t} = G(x_s, s, t)$, where $x_{s \rightarrow t}$ is obtained by ODE solver from $s$ to $t$.

For simplicity, the absolute value notation is omitted in following derivation.
\begin{equation}
\begin{aligned}
    L_{CD} =&{} ||f_\theta(x_{t_{n}}, t_{n}) - f_\theta(x_{t_n \rightarrow t_{n+1}}, t_{n+1})||_2^2 \\
    =&{} c_{skip}(t_{n})x_{t_{n}} + c_{out}(t_{n})F_\theta (x_{t_{n}}, t_{n}) - c_{skip}(t_{n+1})x_{t_n \rightarrow t_{n+1}} - c_{out}(t_{n+1})F_\theta (x_{t_n \rightarrow t_{n+1}}, t_{n+1}) \\
    =&{} \underbrace{[c_{skip}(t_{n})x_{t_{n}} - c_{skip}(t_{n+1})x_{t_n \rightarrow t_{n+1}}]}_{\textbf{term 1}} + \underbrace{c_{out}(t_{n})[F_\theta (x_{t_{n}}, t_{n}) - F_\theta (x_{t_n \rightarrow t_{n+1}}, t_{n+1})]}_{\textbf{term 2}} \\
    &+ \underbrace{[c_{out}(t_{n}) - c_{out}(t_{n+1})]F_\theta(x_{t_n \rightarrow t_{n+1}}, t_{n+1})}_{\textbf{term 3}}
\end{aligned}
\end{equation}
where $x_{t_n \rightarrow t_{n+1}}$ is calculated given ODE solver $G$ as $x_{t_n \rightarrow t_{n+1}} = G(x_{t_n}; t_n, t_{n+1})$.

\textbf{term 1}:
\begin{equation}
    \begin{aligned}
        [c_{skip}(t_{n})x_{t_{n}} - c_{skip}(t_{n+1})x_{t_n \rightarrow t_{n+1}}] 
        =&{} c_{skip}(t_n)x_{t_n} - c_{skip}(t_{n+1}) G(x_{t_n}; t_n, t_{n+1}) \\
        =&{} c_{skip}(t_n)x_{t_n} - c_{skip}(t_{n+1}) (G(x_{t_n}; t_n, t_{n+1}) - x_{t_{n+1}} + x_{t_{n+1}}) \\
        =&{} [c_{skip}(t_n)x_{t_n} - c_{skip}(t_{n+1})x_{t_{n+1}}] + c_{skip}(t_{n+1}) (x_{t_{n+1}} - G(x_{t_{n}};t_{n}, t_{n+1})) \\
        \le&{} c_{skip}(t_n)(\alpha_{t_n} x_\pi + \sigma_{t_n} \epsilon]) - c_{skip}(t_{n+1})(\alpha_{t_{n+1}} x_\pi + \sigma_{t_{n+1}} \epsilon]) \\
        &+ c_{skip}(t_{n+1}) \underbrace{O((t_{n} - t_{n+1})^{p+1})}_{\textbf{local error of Euler soler}} \\
        =&{} (c_{skip}(t_n)\alpha_{t_n} -c_{skip}(t_{n+1})\alpha_{t_{n+1}})x_\pi \\
        &+ [c_{skip}(t_n)\sigma_{t_n} - c_{skip}(t_{n+1})\sigma_{t_{n+1}}]\epsilon \\
        &+ c_{skip}(t_{n+1})O((\triangle t)^{p+1})
    \end{aligned}
\end{equation}
where $0 < \alpha_{t_n}, \sigma_{t_n} < 1$. When $t \ge 30$, $c_{skip}(t)$ is monotonically decreasing, and $0 < c_{skip}(t)\alpha_{t} \le 10^{-7}$. Because all experiments were conducted under FP16 configuration, $c_{skip}(t)\alpha_{t}, c_{skip}(t)\sigma_{t} \approx 0$. Beside, $t$ exceeds 100 in the vast majority of cases. term 1 can be simplified as:
\begin{equation}
    [c_{skip}(t_{n})x_{t_{n}} - c_{skip}(t_{n+1})x_{t_n \rightarrow t_{n+1}}] = m + c_{skip}(t_{n+1})O((\triangle t)^{p+1})
\end{equation}
where $m$ is a value so small that it is effectively negligible in devices.

\textbf{term 2}: We use the DPM-solver as ODE solver $G(x_s, s, t):$
\begin{equation}
    \begin{aligned}
[F_\theta (x_{t_{n}}, t_{n}) - F_\theta (x_{t_n \rightarrow t_{n+1}}, t_{n+1})] 
=&{} \frac{x_{t_n} - \sigma_{t_n} \epsilon_\theta(x_{t_n}, t_n)}{\alpha_{t_n}} - \frac{x_{t_n \rightarrow t_{n+1}} - \sigma_{t_{n+1}}\epsilon_\theta(x_{t_n \rightarrow t_{n+1}, t_{n+1}})}{\alpha_{t_{n+1}}} \\
=&{} \frac{x_{t_n} - \sigma_{t_n} \epsilon_\theta(x_{t_n}, t_n)}{\alpha_{t_n}} \\
&- \frac{\frac{\alpha_{t_{n+1}}}{\alpha_{t_n}}x_{t_n} - \sigma_{t_{n+1}}(e^{-h}-1) \epsilon_\theta(x_{t_n}, t_n) + O(h^2) - \sigma_{t_{n+1}}\epsilon_\theta(x_{t_n \rightarrow t_{n+1}, t_{n+1}})}{\alpha_{t_{n+1}}} \\
=&{} \frac{x_{t_n} - \sigma_{t_n} \epsilon_\theta(x_{t_n}, t_n)}{\alpha_{t_n}} \\
&- \frac{\frac{\alpha_{t_{n+1}}}{\alpha_{t_n}}x_{t_n} - \alpha_{t_{n+1}}(\frac{\sigma_{t_n}}{\alpha_{t_n}} - \frac{\sigma_{t_{n+1}}}{\alpha_{t_{n+1}}}) \epsilon_\theta(x_{t_n}, t_n) + O(h^2) - \sigma_{t_{n+1}}\epsilon_\theta(x_{t_n \rightarrow t_{n+1}, t_{n+1}})}{\alpha_{t_{n+1}}} \\
=&{} \frac{x_{t_n} - \sigma_{t_n} \epsilon_\theta(x_{t_n}, t_n)}{\alpha_{t_n}} \\
&- (\frac{x_{t_n}}{\alpha_{t_n}} - \frac{\sigma_{t_n}}{\alpha_{t_n}}\epsilon_\theta(x_{t_{n}}, t_n) + \frac{\sigma_{t_{n+1}}}{\alpha_{t_{n+1}}}(\epsilon_\theta(x_{t_n}, t_n) - \epsilon_\theta(x_{t_n \rightarrow t_{n+1}, t_{n+1}}))) - \frac{O(h^2)}{\alpha_{t_{n+1}}} \\
=&{} \frac{\sigma_{t_{n+1}}}{\alpha_{t_{n+1}}}(\epsilon_\theta(x_{t_n \rightarrow t_{n+1}, t_{n+1}}) - \epsilon_\theta(x_{t_n}, t_n)) - \frac{O(h^2)}{\alpha_{t_{n+1}}}
\end{aligned}
\end{equation}
Finally, term 2 can be simplified as followed:
\begin{equation}
    c_{out}(t_{n})[F_\theta (x_{t_{n}}, t_{n}) - F_\theta (x_{t_n \rightarrow t_{n+1}}, t_{n+1})] = c_{out}(t_n)\frac{\sigma_{t_{n+1}}}{\alpha_{t_{n+1}}}(\epsilon_\theta(x_{t_n \rightarrow t_{n+1}, t_{n+1}}) - \epsilon_\theta(x_{t_n}, t_n)) - \frac{O(h^2)}{\alpha_{t_{n+1}}}
\end{equation}

\textbf{term 3}: 
\begin{equation}
\begin{aligned}
    [c_{out}(t_{n}) - c_{out}(t_{n+1})]F_\theta(x_{t_n \rightarrow t_{n+1}}, t_{n+1}) 
    =&{}  [c_{out}(t_{n}) - c_{out}(t_{n+1})](\frac{x_{t_n \rightarrow t_{n+1}} - \sigma_{t_{n+1}} \epsilon_\theta(x_{t_n \rightarrow t_{n+1}}, t_{n+1})}{\alpha_{t_{n+1}}}) \\
    \stackrel{(i)}{=}&{} [c_{out}(t_{n}) - c_{out}(t_{n+1})](\frac{x_{t_n} - \sigma_{t_n}\epsilon_\theta(x_{t_n}, t_n)}{\alpha_{t_n}} ) \\
    &+  [c_{out}(t_{n}) - c_{out}(t_{n+1})](\frac{\sigma_{t_{n+1}}}{\alpha_{t_{n+1}}}(\epsilon_\theta(x_{t_n}, t_n) - \epsilon_\theta(x_{t_n \rightarrow t_{n+1}}, t_{n+1}))) \\
    =&{} [c_{out}(t_{n}) - c_{out}(t_{n+1})]x_\pi \\ 
    &+ [c_{out}(t_{n}) - c_{out}(t_{n+1})](\frac{\sigma_{t_n}}{\alpha_{t_n}})(\epsilon - \epsilon_\theta(x_{t_n}, t_n)) \\
    &+  [c_{out}(t_{n}) - c_{out}(t_{n+1})](\frac{\sigma_{t_{n+1}}}{\alpha_{t_{n+1}}}(\epsilon_\theta(x_{t_n}, t_n) - \epsilon_\theta(x_{t_n \rightarrow t_{n+1}}, t_{n+1})))
    \end{aligned}
\end{equation}
where $(i)$ hold according to the DPM-solver.

In our setting, the time interval is 100 and $t \ge 30$. Therefore, $|c_{out}(t_{n}) - c_{out}(t_{n+1})| \le 10^{-7}$. However, in early stage, $\frac{\sigma_{t_{n}}}{\alpha_{t_{n}}} \ge 1$ and $[c_{out}(t_{n}) - c_{out}(t_{n+1})](\frac{\sigma_{t_{n}}}{\alpha_{t_{n}}})$  cannot be neglected. As number of iterations increases, the time $t$ will gradually decrease, and $\frac{\sigma_{t_{n}}}{\alpha_{t_{n}}}  < 1$. At this point,$[c_{out}(t_{n}) - c_{out}(t_{n+1})](\frac{\sigma_{t_{n}}}{\alpha_{t_{n}}})$ can be neglected. Then, term 3 can be formula as:
\begin{equation}
    \begin{aligned}
        [c_{out}(t_{n}) - c_{out}(t_{n+1})]F_\theta(x_{t_n \rightarrow t_{n+1}}, t_{n+1}) 
        =&{} m + [c_{out}(t_{n}) - c_{out}(t_{n+1})](\frac{\sigma_{t_n}}{\alpha_{t_n}})(\epsilon - \epsilon_\theta(x_{t_n}, t_n)) \\
    &+  [c_{out}(t_{n}) - c_{out}(t_{n+1})](\frac{\sigma_{t_{n+1}}}{\alpha_{t_{n+1}}}(\epsilon_\theta(x_{t_n}, t_n) - \epsilon_\theta(x_{t_n \rightarrow t_{n+1}}, t_{n+1}))) 
    \end{aligned}
\end{equation}

Finally, combining the results from term 1, term 2, and term 3, we obtain our CTS loss:
\begin{equation} \label{lcd}
    \begin{aligned}
         L_{CD} =&{} ||f_\theta(x_{t_{n}}, t_{n}) - f_\theta(x_{t_n \rightarrow t_{n+1}}, t_{n+1})||_2^2 \\
         =&{} ||m + c_{skip}(t_{n+1})O((\triangle t)^{p+1}) \\
         &+ c_{out}(t_n)\frac{\sigma_{t_{n+1}}}{\alpha_{t_{n+1}}}(\epsilon_\theta(x_{t_n \rightarrow t_{n+1}, t_{n+1}}) - \epsilon_\theta(x_{t_n}, t_n)) - \frac{O(h^2)}{\alpha_{t_{n+1}}} \\
         &+ m + [c_{out}(t_{n}) - c_{out}(t_{n+1})](\frac{\sigma_{t_n}}{\alpha_{t_n}})(\epsilon - \epsilon_\theta(x_{t_n}, t_n)) \\
        &+[c_{out}(t_{n}) - c_{out}(t_{n+1})](\frac{\sigma_{t_{n+1}}}{\alpha_{t_{n+1}}}(\epsilon_\theta(x_{t_n}, t_n) - \epsilon_\theta(x_{t_n \rightarrow t_{n+1}}, t_{n+1})))||_2^2 \\
        \stackrel{(ii)}{=}&{} ||c_{out}(t_{n+1})(\frac{\sigma_{t_{n+1}}}{\alpha_{t_{n+1}}})(\epsilon_\theta(x_{t_n \rightarrow t_{n+1}}, t_{n+1}) - \epsilon_\theta(x_{t_n}, t_n))||_2^2 + 
        ||[c_{out}(t_{n+1}) - c_{out}(t_{n})](\frac{\sigma_{t_n}}{\alpha_{t_n}})(\epsilon_\theta(x_{t_n}, t_n) - \epsilon)||_2^2 \\
        & - \frac{O(h^2)}{\alpha_{t_{n+1}}} + c_{skip}(t_{n+1})O((\triangle t)^{p+1}) + m \\
        =&{} L_{CTS} - \frac{O(h^2)}{\alpha_{t_{n+1}}} + c_{skip}(t_{n+1})O((\triangle t)^{p+1}) + m
    \end{aligned}
\end{equation}
where $(ii)$ is because we follow DreamFusion\cite{dreamfusion} and omit the U-Net Jacobian term in practice. $\frac{\sigma_{t_n}}{\alpha_{t_n}} > 1$ in early stage, we can not omit the second term.

Therefore, we obtain two conclusions:
\begin{itemize}
    \item $L_{CTS}$ can be expressed as the sum the $L_{CD}$ and two infinitesimal components, along with a term whose magnitude is bounded by $10^{-7}$.
    \begin{equation}
        \begin{aligned}
            L_{CD} &= L_{CTS} + \underbrace{(-\frac{O(h^2)}{\alpha_{t_{n+1}}}) + c_{skip}(t_{n+1})O((\triangle t)^{p+1})}_{\text{infinitesimal components}} + \underbrace{m}_{|m| \le 10^{-7}} \\
            h &= log(\frac{\alpha_{t_n}}{\sigma_{t_n}}) - log(\frac{\alpha_{t_{n+1}}}{\sigma_{t_{n+1}}})
        \end{aligned}
    \end{equation}

    \item $L_{CTS}$ can formula as:
    \begin{equation}
        \begin{aligned}
            L_{CTS} &= ||w_1(\epsilon_\theta(x_{t_n \rightarrow t_{n+1}}, t_{n+1}) - \epsilon_\theta(x_{t_n}, t_n))||_2^2 + 
        ||w_2(\epsilon_\theta(x_{t_n}, t_n) - \epsilon)||_2^2 \\
        w_1 &= c_{out}(t_{n+1})(\frac{\sigma_{t_{n+1}}}{\alpha_{t_{n+1}}}) \\
        w_2 &= [c_{out}(t_{n+1}) - c_{out}(t_{n})](\frac{\sigma_{t_n}}{\alpha_{t_n}})
        \end{aligned}
    \end{equation}
    
\end{itemize}
The proof is completed.

\end{proof}

\end{theorem}

\begin{theorem} \label{the2}
Assume that the pre-trained noise predictor $\epsilon_\theta(\cdot;\cdot)$ satisfies the Lipschitz condition. Define $\triangle := sup|t_1 - t_2)|$. For any given camera pose $\pi$, if convergence is achieved according to $L_{CTS}$, then there exists a corresponding real image $x_0 \sim p_{data}(x)$ such that:
\begin{equation}
    ||x_\pi - x_0||_2 = O(\triangle)
\end{equation}
where $x_\pi = g(\theta, \pi)$ denotes the rendered image for pose $\pi$.

We offer two ways of proof. For the \textbf{Proof 1}, we directly utilize the CTS loss. For the \textbf{Proof 2}, we make use of Theorem \ref{the1}.

\textbf{Proof 1:}
\begin{proof*}
    Given $L_{CTS}(\xi) = 0$, for any $t, s$, we have $\epsilon({x_{s \rightarrow t}}, t) = \epsilon_\theta(x_s, s)$ and $T \ge t_n \ge t_{n-1} \ge 0$.
    Assume $G$ is DPM solver and follow the first-order definition of DPM-Solver, given $e$, we have:
    \begin{equation}
        \begin{aligned}
        G(x_{s \rightarrow t}, t, e) =&{} \frac{\alpha_e}{\alpha_t} x_{s \rightarrow t} - \sigma_e (e^{-h} - 1) \epsilon_\theta(x_{s \rightarrow t}, t) \\
        =&{} \frac{\alpha_e}{\alpha_t}x_{s \rightarrow t} - \alpha_e(\frac{\sigma_t}{\alpha_t} - \frac{\sigma_e}{\alpha_e})\epsilon_\theta(x_{s \rightarrow t}, t) \\ 
        =&{} \alpha_e \frac{x_{s \rightarrow t} - \sigma_t \epsilon_\theta(x_{s \rightarrow t}, t)}{\alpha_t} + \sigma_e \epsilon_\theta(x_{s \rightarrow t}, t) \\
        =&{} \alpha_e \frac{\alpha_t(\frac{x_s - \sigma_s \epsilon_\theta(x_s, s)}{\alpha_s}) + \sigma_t \epsilon(x_s, s) - \sigma_t \epsilon_\theta(x_{s \rightarrow t}, t) + O(h^2)}{\alpha_t} + \sigma_e \epsilon_\theta(x_{s \rightarrow t}, t) \\
        \stackrel{(iii)}{=}&{} \alpha_e (\frac{x_s - \sigma_s \epsilon_\theta(x_s, s)}{\alpha_s}) + \sigma_e \epsilon_\theta(x_{s}, s) + L_1 O(h^2) \\
        =&{} G(x_s, s, e) + L_1 O(h^2)
        \end{aligned}
    \end{equation}
where $(iii)$ hold according to the $\epsilon_{\theta}({x_{s \rightarrow t}}, t) = \epsilon_\theta(x_s, s)$.
When we set $e = 0$, the $G(x_s, s, 0)$ can be treated as diffusion model $D$ that directly predicts the original image $x_0$. We define $D(x_s, s) = G(x_s, s, 0)$ and $D(x_{s \rightarrow t}, t) = D(x_s, s) + L_1 O(h^2)$. 

Let $e_n$ represent the error at $t_n$, which is defined as:
\begin{equation}
    e_n := D(x_{t_n}, t_n) - x_0.
\end{equation}
We can derive the error at $t_{n+1}$ given the error at $t_n$:
\begin{equation}
    \begin{aligned}
        e_n =&{} D(x_{t_n}, t_n) - x_0 \\
        =&{} D(x_{t_n}, t_n) - D(x_{t_{n-1} \rightarrow t_n}, t_n) + D(x_{t_{n-1} \rightarrow t_n}, t_n) - x_0 \\
        =&{} \frac{x_{t_n} - \sigma_{t_n}
        \epsilon_\theta(x_{t_n}, t_n)}{\alpha_{t_n}} - \frac{x_{t_{n-1} \rightarrow t_n} - \sigma_{t_n}
        \epsilon_\theta(x_{t_{n-1} \rightarrow t_n}, t_n)}{\alpha_{t_n}} \\ 
        &+ D(x_{t_{n-1}}, t_{n-1}) - x_0 + L_1 O((t_n - t_{n-1})^2) \\
        =&{} \frac{1}{\alpha_{t_n}}(x_{t_n} - x_{t_{n-1} \rightarrow t_n}) + \frac{\sigma_{t_n}}{\alpha_{t_n}}(\epsilon_\theta(x_{t_{n-1} \rightarrow t_n}, t_n) - \epsilon_\theta(x_{t_n}, t_n)) + e_{n-1} + L_1 O((t_n - t_{n-1})^2) \\
    \end{aligned}
\end{equation}
According to the Lipschitz condition, we can further derive:
\begin{equation}
    \begin{aligned}
        ||e_n|| \le&{} \frac{1}{\alpha_{t_n}}||x_{t_n} - x_{t_{n-1}\rightarrow t_n}|| + \frac{\sigma_{t_n}}{\alpha_{t_n}}||\epsilon_\theta(x_{t_{n-1} \rightarrow t_n}, t_n) - \epsilon_\theta(x_{t_n}, t_n)|| + ||e_{n-1}|| + L_1 O((t_n - t_{n-1})^2) \\
        \stackrel{(iv)}{\le}&{} \frac{1}{\alpha_{t_n}} O((t_n - t_{n-1})^2) + \frac{\sigma_{t_n}}{\alpha_{t_n}}L||x_{t_{n-1} \rightarrow t_n} - x_{t_n}|| + ||e_{n-1}|| + L_1 O((t_n - t_{n-1})^2) \\
        \stackrel{(v)}{\le}&{} ||e_{n-1}|| + KO((t_n - t_{n-1})^2)
    \end{aligned}
\end{equation}
where $(iv), (v)$ hold according to the local error of Euler solver and Lipschitz condition of $\epsilon_\theta$. Therefore, we can derive the error recursively:
\begin{equation}
    \begin{aligned}
        ||e_T|| \le&{} K\sum_{i=1}^{N-1}O((t_i - t_{i-1})^2) \\ 
        \le&{} K\sum_{i=1}^{N-1}(t_i - t_{i-1})O(\triangle) \\
        \le&{} KO(\triangle)(T-\xi)
    \end{aligned}
\end{equation}
This shows that CTS is capable of achieving the same accuracy as multi-step approaches in a single-step generative frame
work, thus demonstrating its efficiency and broad applicability for optimization-based generation. 
The proof is completed.
\end{proof*}

\textbf{Proof 2:}
\begin{proof}
    According to theorem \ref{the1} and CM\cite{cm}, the CTS loss is equivalent to consistency loss and the consistency function satisfies the Lipschitz condition. Given $L_{CTS}(\xi) = 0$, we have $L_{CD}(\xi) \le O((\triangle t)^2)$ and $f_\theta (x_{s \rightarrow t}, t) \le f_\theta(x_s, s) + O((\triangle t)^2)$. And $T \ge t_n \ge t_{n-1} \ge 0$.
\begin{equation}
    \begin{aligned}
        e_{n} =&{} f_\theta(x_{t_n}, t_n) - x_0 \\
        =&{} f_\theta (x_{t_n}, t_n) - f_\theta (x_{t_{n-1} \rightarrow x_{n}}, t_n) + f_\theta (x_{t_{n-1} \rightarrow x_{n}}, t_n) - x_0 \\ 
        \le&{} f_\theta(x_{t_n}, t_n) - f_\theta (x_{t_{n-1} \rightarrow x_{n}}, t_n) + f_\theta(x_{t_{n-1}}, t_{n-1}) - x_0 + O((\triangle t)^2)
    \end{aligned}
\end{equation}
According to the Lipschitz condition, we can further derive:
\begin{equation}
    \begin{aligned}
        ||e_n|| \le&{} ||f_\theta(x_{t_n}, t_n) - f_\theta(x_{t_{n-1} \rightarrow x_n}, x_n)|| + ||e_{n-1}|| + O((\triangle t)^2) \\
        \stackrel{(vi)}{\le}&{} L ||x_{t_n} - x_{t_{n-1} \rightarrow x_n}||  + ||e_{n-1}|| + O((\triangle t)^2) \\
        \stackrel{(vii)}{\le}&{} L(O(t_n - t_{n+1})^2) + ||e_{n-1}|| + O((\triangle t)^2) \\
        \le&{} (L+1)O((t_n - t_{n+1})^2) + ||e_{n-1}|| \\
        =&{} ||e_{n-1}|| + O((t_n - t_{n+1})^2)
    \end{aligned}
\end{equation}
where $(vi)$ and $(vii)$ hold according to the Lipschitz condition and local error of Euler solver respectively. Therefor, we can drive the error recursively:
\begin{equation}
    \begin{aligned}
        ||e_T|| \le&{} \sum_{i=1}^{N-1} O((t_i - t_{i-1})^2) \\
        \le&{} \sum_{i=1}^{N-1} (t_i - t_{i-1}) O((\triangle)) \\
        \le&{} O((\triangle))(T - \xi)
    \end{aligned}
\end{equation}
The proof is completed.
\end{proof}

\end{theorem}
\begin{corollary}
Based on Theorem. \ref{the1}, we can derive two conclusions: 
\begin{itemize}
    \item Existing methods\cite{DreamLCM, vividdreamer} which incorporate Latent Consistency Model for 3D generation can be amalgamated into our framework.
    \item Consistency function\cite{cm} can ensures both self-consistency and cross-consistency.
\end{itemize}

Consistent3D\cite{consistent3d} and CCD\cite{CCD} drive the inspiration from Consistency model\cite{cm}, however, they are still trained based on Stable Diffusion-2-1\cite{ddpm}. Hence, we merely consider DreamLCM\cite{DreamLCM} and Vividdreamer\cite{vividdreamer} that based on LCM\cite{lcm}.

\begin{proof}
    DreamLCM\cite{DreamLCM}: 
    \begin{equation}
        L_{DreamLCM} = \mathbb{E}_{t,c}[w(t)(\epsilon_\theta(x_t, t) -\epsilon)\frac{\partial g(\theta, c)}{\partial \theta}]
    \end{equation}
    where $x_t$ is obtained by Euler Solver from $x_s$ and $x_s = \alpha_s x_\pi + \sigma_s \epsilon$, $\epsilon \in N(0, I)$.

    Vividreamer\cite{vividdreamer}:
    \begin{equation}
    \begin{aligned}
        L_{vividreamer} =&{} \mathbb{E}_{t,c}[||w(t)(x_\pi - f(x_s, s))||_2^2] \\
        \end{aligned}
    \end{equation}
    where $x_s = \alpha_sx_\pi + \sigma_s \epsilon$. 

For DreamLCM\cite{DreamLCM}, according to eq. \ref{lcd}, $L_{CD}$ can express as:
\begin{equation}
    \begin{aligned}
        L_{CD} =&{} ||f_\theta(x_{t_{n}}, t_{n}) - f_\theta(x_{t_n \rightarrow t_{n+1}}, t_{n+1})||_2^2 \\
         =&{} ||m + c_{skip}(t_{n+1})O((\triangle t)^{p+1}) \\
         &+ c_{out}(t_n)\frac{\sigma_{t_{n+1}}}{\alpha_{t_{n+1}}}(\epsilon_\theta(x_{t_n \rightarrow t_{n+1}, t_{n+1}}) - \epsilon_\theta(x_{t_n}, t_n)) - \frac{O(h^2)}{\alpha_{t_{n+1}}} \\
         &+ m + [c_{out}(t_{n}) - c_{out}(t_{n+1})](\frac{\sigma_{t_n}}{\alpha_{t_n}})(\epsilon - \epsilon_\theta(x_{t_n}, t_n)) \\
        &+[c_{out}(t_{n}) - c_{out}(t_{n+1})](\frac{\sigma_{t_{n+1}}}{\alpha_{t_{n+1}}}(\epsilon_\theta(x_{t_n}, t_n) - \epsilon_\theta(x_{t_n \rightarrow t_{n+1}}, t_{n+1})))||_2^2 \\
        =&{}  ||m + c_{skip}(t_{n+1})O((\triangle t)^{p+1}) \\
         &+ c_{out}(t_{n+1})\frac{\sigma_{t_{n+1}}}{\alpha_{t_{n+1}}}(\epsilon_\theta(x_{t_n \rightarrow t_{n+1}, t_{n+1}}) - \epsilon_\theta(x_{t_n}, t_n)) - \frac{O(h^2)}{\alpha_{t_{n+1}}} \\
         &+ m + [c_{out}(t_{n}) - c_{out}(t_{n+1})](\frac{\sigma_{t_n}}{\alpha_{t_n}})(\epsilon - \epsilon_\theta(x_{t_n \rightarrow t_{n+1}, t_{n+1}}) + \epsilon_\theta(x_{t_n \rightarrow t_{n+1}, t_{n+1}}) - \epsilon_\theta(x_{t_n}, t_n))||_2^2 \\ 
         =&{} m + c_{skip}(t_{n+1})O((\triangle t)^{p+1}) \\
         &+ [c_{out}(t_{n+1})\frac{\sigma_{t_{n+1}}}{\alpha_{t_{n+1}}} + [c_{out}(t_{n}) - c_{out}(t_{n+1})](\frac{\sigma_{t_n}}{\alpha_{t_n}})](\epsilon_\theta(x_{t_n \rightarrow t_{n+1}, t_{n+1}}) - \epsilon_\theta(x_{t_n}, t_n)) \\
         &+ [c_{out}(t_{n}) - c_{out}(t_{n+1})](\frac{\sigma_{t_n}}{\alpha_{t_n}})\underbrace{ (\epsilon - \epsilon_\theta(x_{t_n \rightarrow t_{n+1}, t_{n+1}})}_{\text{loss of DreamLCM}}
    \end{aligned}
\end{equation}
DreamLCM assumes that the denoising process of LCM follows a smooth PF-ODE trajectory with a small slope. This assumption allows us to omit the term of $(\epsilon_\theta(x_{t_n \rightarrow t_{n+1}, t_{n+1}}) - \epsilon_\theta(x_{t_n}, t_n))$, therefore, in this assumption, $L_{DreamLCM}$ is a special case of $L_{CTS}$.

For Vividdreamer\cite{vividdreamer}, we have:
\begin{equation}
    \begin{aligned}
        L_{vividdreamer} =&{} ||x_\pi - f(x_s, s)||_2^2 \\
        =&{} ||x_\pi - (c_{in}(s)x_\pi + c_{skip}(s)\frac{x_s - \sigma_s \epsilon_\theta (x_s, s)}{\alpha_s})||_2^2 \\
        =&{} || (1 - c_{skip}(s))x_\pi - c_{in}x_s + c_{skip}(s)x_\pi - c_{skip}(s) \frac{x_s - \sigma_s \epsilon_\theta (x_s, s)}{\alpha_s} ||_2^2 \\
        =&{} ||(1 - c_{skip}(s))x_\pi - c_{in}(\alpha_s x_\pi + \sigma_s \epsilon) + c_{skip}(s)\frac{\alpha_s x_\pi - \alpha_s x_\pi - \sigma_s (\epsilon - \epsilon_\theta(x_s, s))}{\alpha_s}||_2^2 \\
        =&{} ||(1 - c_{skip}(s))x_\pi - c_{in}(s)(\alpha_s x_\pi + \sigma_s \epsilon) + c_{skip}(s)\frac{\sigma_s}{\alpha_s}(\epsilon - \epsilon_\theta(x_s, s))||_2^2 \\
        =&{} ||[1 - c_{skip}(s) - c_{in}(s)\alpha_s]x_\pi - c_{in}(s)\sigma_s \epsilon + c_{skip}(s)\frac{\sigma_s}{\alpha_s}(\epsilon - \epsilon_\theta(x_s, s))||_2^2 || \\
        \stackrel{(viii)}{\approx} &{} c_{skip}(s)\frac{\sigma_s}{\alpha_s}(\epsilon - \epsilon_\theta(x_s, s))||_2^2 
    \end{aligned}
\end{equation}
where $(viii)$ hold according to the $c_{skip}(s) + c_{in}(s) \approx 1$ and $\mathbb{E}[\epsilon] = 0$. Therefore, Vividdreamer is a special case of our CTS loss.

According to the definition of self-consistency and cross-consistency(SDS) in previous methods\cite{consistent3d, CCD}, we can observe that $L_{CTS}$ simultaneously ensures two types of consistency:
\begin{equation}
    \begin{aligned}
        L_{CTS} =&{} \underbrace{||c_{out}(t_{n+1})(\frac{\sigma_{t_{n+1}}}{\alpha_{t_{n+1}}})(\epsilon_\theta(x_{t_n \rightarrow t_{n+1}}, t_{n+1}) - \epsilon_\theta(x_{t_n}, t_n))||_2^2}_{\textbf{self-consistency term}} \\
        &+ \underbrace{||[c_{out}(t_{n+1}) - c_{out}(t_{n})](\frac{\sigma_{t_n}}{\alpha_{t_n}})(\epsilon_\theta(x_{t_n}, t_n) - \epsilon)||_2^2}_{\textbf{cross-consistency term(SDS loss)}}
    \end{aligned}
\end{equation}
The proof is complete.
\end{proof}

\end{corollary}

\begin{figure}
    \centering
    \includegraphics[width=1.0\linewidth]{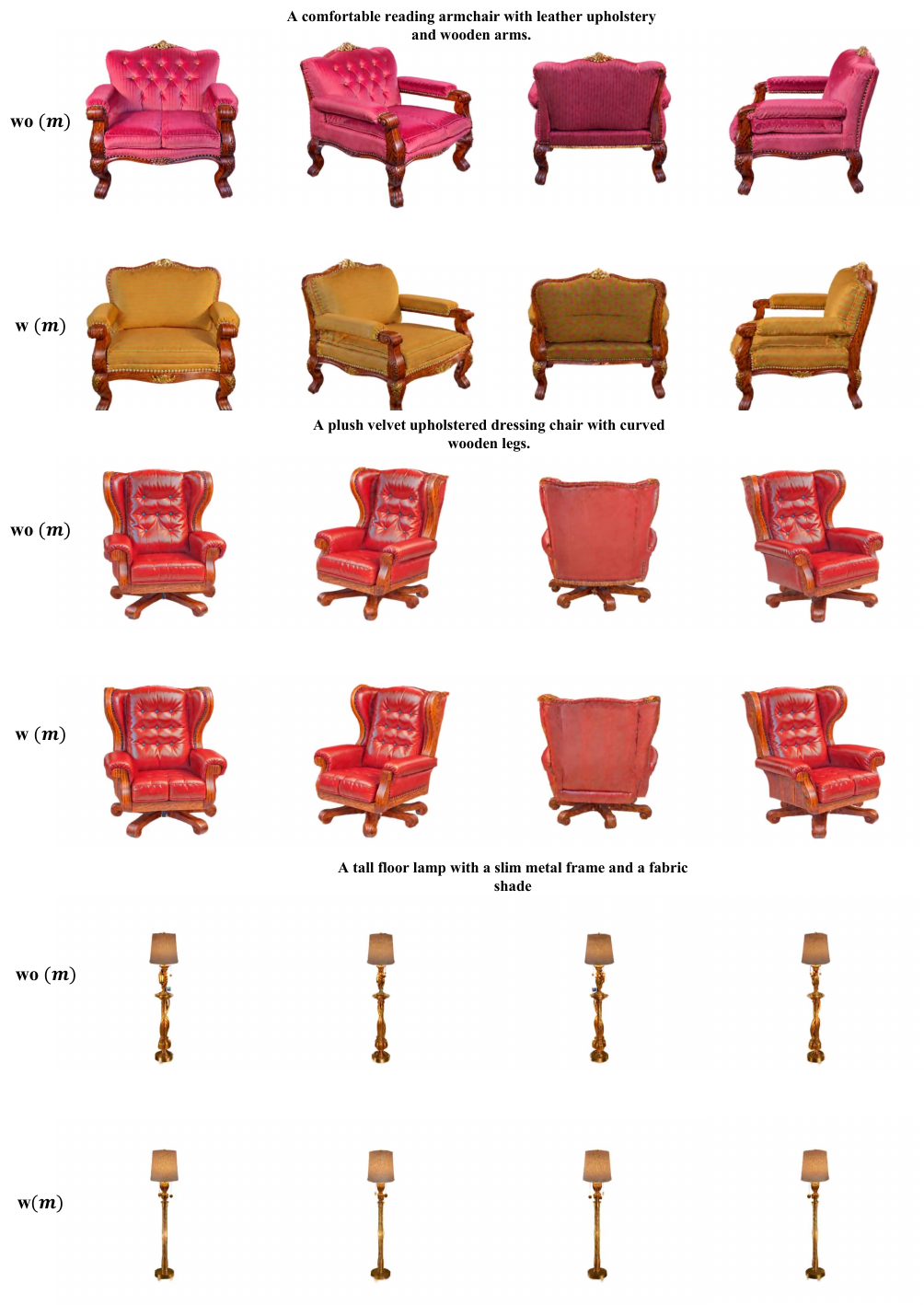}
    \caption{Ablation Study of $m$ in Object Generation.}
    \label{fig:ablation_m_obj}
\end{figure}

\begin{figure}
    \centering
    \includegraphics[width=1.0\linewidth]{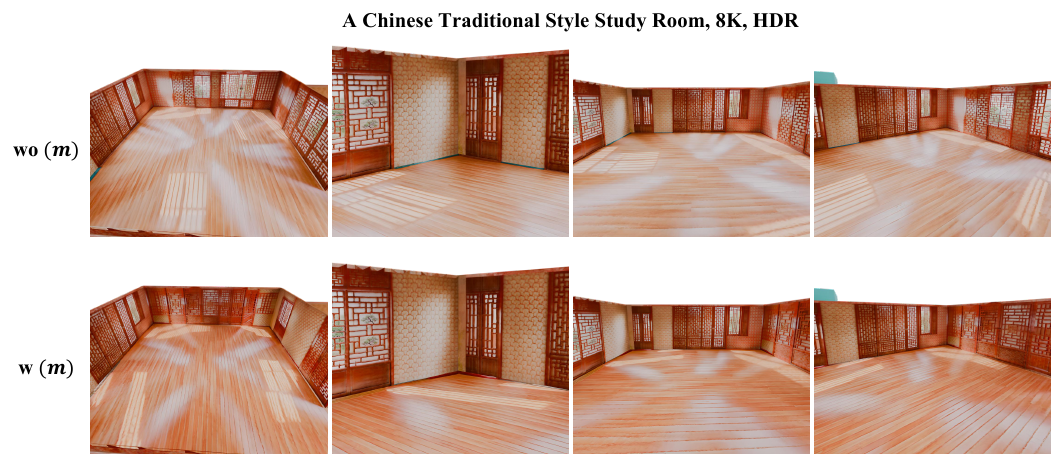}
    \caption{Ablation Study of $m$ in Scene Generation. After training, we distill the texture field into a texture map and import the environment into Blender\cite{blender} for visualization.}
    \label{fig:ablation_m_scene}
\end{figure}

\subsubsection{Ablation Study for $m$}
\textbf{It is worth noting that our generation process exhibits inherent randomness, meaning identical prompts may not consistently produce the exact same objects or scenes. Therefore, we primarily compare the overall texture trends and fine details.} 
Next, we conduct ablation studies on $m$ separately for objects and scenes synthesis. The experimental results demonstrate that the $m$ has negligible impact on the results, whether synthesizing objects or scenes. 

\paragraph{Scenes}
As shown in Figure \ref{fig:ablation_m_scene}. Despite slight color differences, the general texture patterns and detailed features are identical.

\paragraph{Objects}
As shown in Figure \ref{fig:ablation_m_obj}, the overall shape and fine details of the objects remain consistent.

\subsection{Pipeline} \label{Pipeline}
Due to space limitations in the regular paper, we omit detailed descriptions of certain technical implementations. In this section, we provide a comprehensive account of our contributions and present the overall pipeline in detail.

Our Contributions for Indoor Scene Generation:
\begin{itemize}
    \item Layout Generation:
        \begin{itemize}
            \item \textbf{Iteractive Programmatic Verification Mechanism:} Ensure that no overlapping occurs between furniture.
            \item \textbf{Cluster-based Orientation Assignment Strategy:} Ensure accuracy orientation generation.
        \end{itemize}
    \item Object Generation:
        \begin{itemize}
            \item \textbf{Consistency Trajectory Sampling(CTS):} Realistic furniture generation.
            \item \textbf{CTS without noise:} Efficiently generating furniture.
            \item \textbf{theorem analysis}: 2 theorem and 1 corollary.
        \end{itemize}
    \item Environment Optimization:
        \begin{itemize}
            \item \textbf{multi-resolution texture field and normal-aware decoder:} high-resolution, style consistency, and rich texture generation
            \item \textbf{Optimization via CTS without noise:} Realistic texture generation
            \item \textbf{zigzag adaptive camera trajectory:} Multi-scale Scene Generation
            \item \textbf{several techniques:} Improve numerical stability
        \end{itemize}
    \item Physical edition: Integrate physical simulation to ensure the physical plausible editing. We propose two way to perform physical plausible editing.
        \begin{itemize}
            \item \textbf{Employing bounding boxes as proxies to conduct physics-based simulations}.
            \item \textbf{Extracting meshes as proxies for performing physics-based simulations}.
        \end{itemize}
\end{itemize}
Next, we present a detailed account of our contributions.

\begin{figure}
    \centering
    \includegraphics[width=1.0\linewidth]{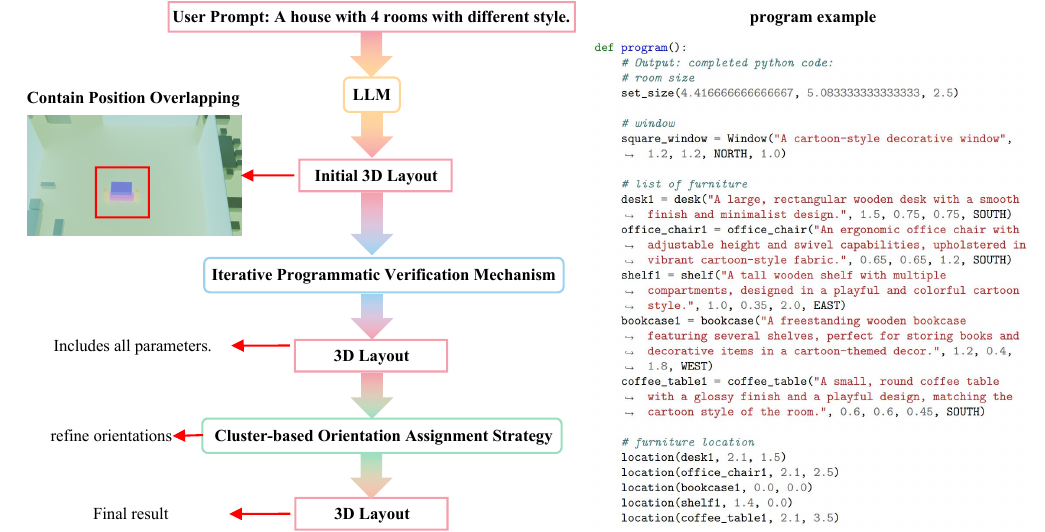}
    \caption{Layout pipeline and code example.}
    \label{fig:layout_pipeline}
\end{figure}

\begin{figure}
    \centering
    \includegraphics[width=1.0\linewidth]{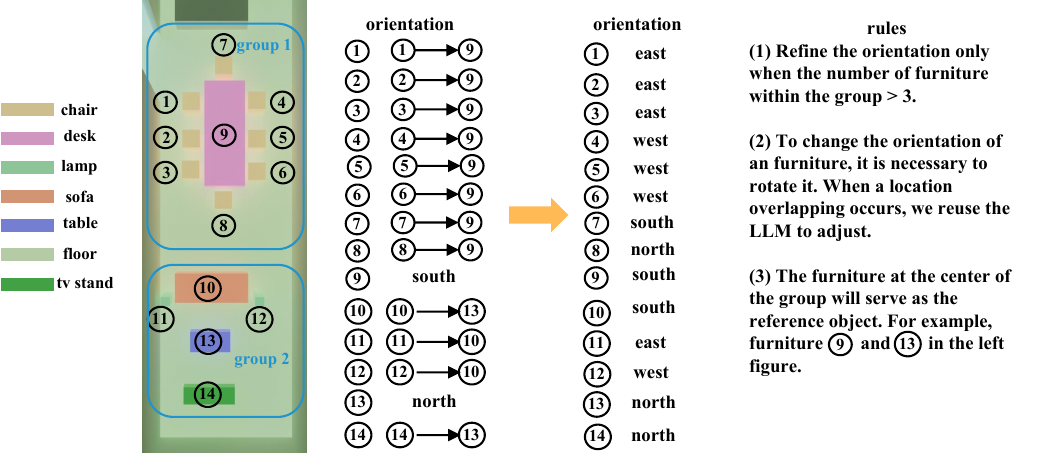}
    \caption{Cluster-based Orientation Assignment.}
    \label{fig:orientation_assignment}
\end{figure}

\subsubsection{Layout Generation}
Layout Generation consists of four key functions as shown in Figure \ref{fig:layout_pipeline}:
\begin{itemize}
    \item User offers a description of the house.
    \item LLM converts the textural description into 3D layout.
        \begin{itemize}
            \item At this point, it already contains initial layout information, including the floorplan, descriptions, positions, and scales of furniture.
            \item However, the initial layout is typically obtained through a single query to the LLM, which often results in overlapping furniture placements.
            \item Next, Iterative Programmatic Verification Mechanism is proposed to address overlapping.
        \end{itemize}
    \item Iterative Programmatic Verification Mechanism
        \begin{itemize}
            \item Convert the 3D layout into programs and execute it to check for conflicts.
            \item Iteratively feed the error and program into the LLM for parameter refinement.
            \item The previous methods\cite{Anyhome, holodeck} generate all the layout parameters in a single query to the LLM. However, when dealing with complex scenes containing numerous furniture, the LLM often struggles to account for all inter-furniture relationships, and it is challenging to generate accurate orientations through a single query.
            \item Next, Cluster-based Orientation Assignment is proposed to generate accuracy orientation.
        \end{itemize}
    \item Cluster-based Orientation Assignment Strategy
        \begin{itemize}
            \item We first partition the furniture into several groups.
            \item Within each group, object orientations are determined not by cardinal directions, but through inter-object spatial relations. In this way, LLM is only required to account for spatial relationships within the current group of furniture.
            \item As shown in Figure \ref{fig:orientation_assignment}, we first select a reference object from within the group to prevent circular argument. Then we generate the orientation of each furniture relative to the others, and finally convert it into cardinal directions.
        \end{itemize}
    \item Convert final program to 3D layout.
\end{itemize}
Our framework can generate more furniture with accurate spatial orientations. \textbf{On average, a single room contains approximately 8 pieces of furniture, while a house with 4 rooms averages around 40 pieces}. In Sec. \ref{ablation_experiments} and Sec. \ref{additional_experiments}, we will compare our method with other approaches and conduct ablation studies to demonstrate its superiority.

\subsubsection{Object Generation}
\paragraph{Motivation.} 
Following Consistent3D\cite{consistent3d}, given a ODE sampling process for a 3D model $\theta$:
\begin{equation}
    \begin{aligned}
        d \theta = - \dot{\sigma}_t \sigma_t \nabla \log p_t(\theta) dt,
    \end{aligned}
\end{equation}
where is randomly initialized according to a certain distribution. And Following DreamFusion\cite{dreamfusion} and SJC\cite{SJC}, the 3D score function $\nabla_\theta \log p_t(\theta)$ from the 2D score function using the chain rule:
\begin{equation} \label{ode_3d}    
\begin{aligned}
        \nabla_\theta \log p_t(\theta) = \mathbb{E}_\pi [\nabla_{x_\pi} \log p_t(x_\pi) \frac{\partial x_\pi}{\partial \theta}],
    \end{aligned}
\end{equation}
where the $x_\pi$ is rendered image, and the 2D score function $\nabla_x \log p_t(x)$ can be estimated as $\nabla_x \log p_t(x) = \frac{D_\phi (x, t) - x}{\sigma_t^2}$ by a pre-trained diffusion model $D_\phi (x, t)$. Therefore, the key to generating a satisfactory 3D model is to accurately perform the 3D ODE sampling in Eq. \ref{ode_3d} using the pre-trained diffusion model. Consistent3D\cite{consistent3d} proposes Consistency Distillation Sampling(CDS) loss to distilling the deterministic prior into the 3D model:
\begin{equation}
    \begin{aligned}
        \mathbb{E}_\pi [\lambda(t)||D_\phi(x_t, t, y) - sg(D_\phi(x_{t \rightarrow s}, s, y))||_2^2],
    \end{aligned}
\end{equation}
where $sg(\cdot)$ is a stop-gradient operator, $ t > s$. They believe that less-noisy sample lies on the 3D model's target ODE trajectory, and are used to guide more-noisy sample. CCD\cite{CCD} builds upon Consistent3D by introducing Conditional Guidance loss and Compact Consistency loss to enhance the consistency. CCD follows Consistent3D and still treats less-noisy sample as golden to guide the more-noisy sample(The stop-gradient operator is not mentioned in the CCD paper, but it is implemented in the CCD code: \url{https://github.com/LMozart/ECCV2024-GCS-BEG.git}). However, less-noisy samples may not necessarily lie on the target ODE trajectory. Treating these as golden for guiding more-noisy samples could introduce artifacts and inconsistencies in the model. Another line of works\cite{Dreamscene, Luciddreamer} also improve the consistency and match two adjacent samples. DreamLCM\cite{DreamLCM} and Vividdreamer\cite{vividdreamer} directly apply Consistency model\cite{lcm} as based model for generating. However, the former is based on the smoothness assumption, while the latter employs a consistency model\cite{lcm} as the based model to enhance the SDS loss\cite{dreamfusion}. Therefore,  these models only adopt the concept of consistency model\cite{consistent3d, CCD} or employ the consistency model as based model\cite{DreamLCM, vividdreamer}, but fail to conduct in-depth research on their interrelationships. 

In contrast, our CTS loss directly originates from the consistency model\cite{cm}, is derived from the consistency function\cite{cm}, and explicitly reveals the relationship between the two. 

Our CTS loss:
\begin{equation}
    \begin{aligned}
        L_{CTS} &= \mathbb{E}[||w_1(t)(\epsilon_\theta(x_{s \rightarrow t}, t, y) - \epsilon_\theta(x_{s}, s, , \emptyset))||_2^2 + 
    ||w_2(s, t)(\epsilon_\theta(x_{s}, s, \emptyset) - \epsilon)||_2^2] \\
    w_1(t) &= c_{out}(t)(\frac{\sigma_{t}}{\alpha_{t}}) \\
    w_2(s, t) &= [c_{out}(t) - c_{out}(s)](\frac{\sigma_{s}}{\alpha_{s}}),
    \end{aligned}
\end{equation}
We can also apply consistency loss\cite{cm}:
\begin{equation}
    \begin{aligned}
        \mathbb{E}[||f(x_s, s, y) - f(x_{s \rightarrow t}, t, y)||_2^2]
    \end{aligned}
\end{equation}
However, this formulation is not conducive to integrating our method with other approaches\cite{CFG, prep_neg} that act on the predicted noise $\epsilon_\theta(\cdot;\cdot)$. Our CTS loss effectively resolves these issues.

\paragraph{Algorithm.} It is worth noting that $w_2(s, t)$ tends to zero as $t$ decreases. Hence we remove the second term when $t \le 500$. Additionally, we employ $L_{scale}$, $L_{layout}$, and $L_{normal}$ for detail. Neither A nor B is our contribution; thus, we will not conduct ablation studies on them in the subsequent ablation experiments. 

\begin{equation}
    \begin{aligned}
        L_{scale} =&{} \frac{1}{N} \sum_{i=1}^N \max(s_i) \\
        L_{normal} =&{} \frac{1}{M} \sum_{i=1}^M w_i (1 - n_i^T \tilde{n}_i) \\
        L_{Layout} =&{} d^x(G_x, x_i, h_i) + d^y(G_y, y_i, w_i) + d^z(G_z, z_i, l_i) \\
        d^x(G_x, x_i, h_i) =&{} ||\min{(G_x)} - (x_i - \frac{h_i}{2})||_2^2 + || \max{(G_x)} - (x_i + \frac{h_i}{2})||_2^2    \\
        w_i =&{} \alpha_i \prod_{j=1}^{i} (1 - \alpha_i),
    \end{aligned}
\end{equation}
$\alpha_i$ is the pixel translucency determined by the
opacity of the $i-$th Gaussian kernel and the pixel’s position. $s_i \in \mathbb{R}^3$ is scale of $i-$th Gaussian. $N$ is the number of Gaussian, $M$ is the number of pixel. Different from Gala3D\cite{gala3d}, while Gala3D restricts objects to lie strictly within bounding boxes, our $L_{layout}$ optimizes for tight alignment with box boundaries, enabling direct physics simulation using the bounding boxes. For more clarity, we summarize our entire furniture generation procedure with the proposed CTS in Algorithm \ref{alg:scenelcm_object}.

\begin{algorithm}
\caption{SceneLCM for Objects}\label{alg:scenelcm_object}
\begin{algorithmic}
\Require 3D model parameters $\theta$, training iteration $n$, latent consistency model network $\phi$ denoising timestep from $N_{min}$ to $N_{max}$, text prompt $y$, fixed noise $\epsilon$,  warm up timestep $T_{warm\_up}$, warm up step $N_{warm\_up}$, and timestep $t_{cut}$. DPM-Solver $G(\cdot;\cdot, \cdot)$
\For{$i = 1,2,\ldots,N$}
\State $r = 1 - \min(i / N_{warm\_up}, 1)$
\State $T_{min} \gets int(N_{min} + r * N_{warm\_up})$, $T_{max} \gets int(N_{max} + r * N_{warm\_up})$
\State camera pose $c$, $x_\pi = g(\theta, c)$
\State $t$ sample from $[T_{min}, T_{max}]$
\State $s$ sample from $[t - 2 * t_{cut}, t - t_{cut}]$ \Comment{Obtain timestep $s$ and $t$.}

\If {$i \le T_{warm\_up}$}
\State $x_s \gets \alpha_s x_\pi + \sigma_s \epsilon$ \Comment{add noise by DDPM.}
\Else
\State $x_s \gets x_\pi$
\EndIf
\State $x_{s \rightarrow t} \gets G(x_s; s, t)$ \Comment{Obtain $x_{s \rightarrow t}$ by Euler Solver.}
\If {$i \le T_{warm\_up}$}
\State calculate CTS loss
\State $\nabla_\theta L_{CTS} \gets w_2(s, t)(\epsilon_\phi(x_s, s, y) - \epsilon)\frac{\partial g(\theta, c)}{\partial \theta}$
\Else
\State $\nabla_\theta L_{CTS} \gets 0$
\EndIf

\State $\nabla_\theta L_{CTS} \gets \nabla_\theta L_{CTS} + w_1(t)(\epsilon_\phi(x_t, t, y) - \epsilon_\phi(x_s, s, \emptyset))\frac{\partial g(\theta, c)}{\partial \theta}$ 
\State $\theta \gets \theta + \eta \nabla_\theta L_{CTS}$\Comment{Besides the CTS loss, we also have losses $L_{scale}$, $L_{layout}$, and $L_{normal}$, which are omitted here for simplicity.}
\EndFor
\end{algorithmic}
\end{algorithm}

\paragraph{Noise Removal.}
The generation process progresses through two sequential stages over decreasing time $t$: semantic generation and detail generation. Once 3DGS developed distinct semantic signal which align with the text prompts, adding noise drives the model to alter details for text alignment. However, as most text lack high-frequency specific, e.g., "An iron man with while hair. 8K HDR", the focus should be on enhancing detail generation rather than forces alignment. Therefore, we eliminate the noise and set $x_s = x_\pi$ as shown in Algorithm \ref{alg:scenelcm_object}, to allow the model to concentrate on detail generation and improve the training efficiency. In Sec. \ref{ablation_experiments}, we conduct ablation studies to verify both the generation efficiency and the visual details of the generated results.

\textbf{Noise removal primarily impacts processing speed at the object level, but at the scene level, this technique becomes crucial and significantly affects the generation quality. We will verify this point in subsequent experiments.}

\subsubsection{Environment Generation}
Five key components:
\begin{itemize}
    \item Multi-resolution Texture Field
    \item Normal-aware Decoder
    \item Optimization via CTS loss
    \item Zigzag adaptive camera trajectory
    \item Several techniques
\end{itemize}

\begin{figure}
    \centering
    \includegraphics[width=1.0\linewidth]{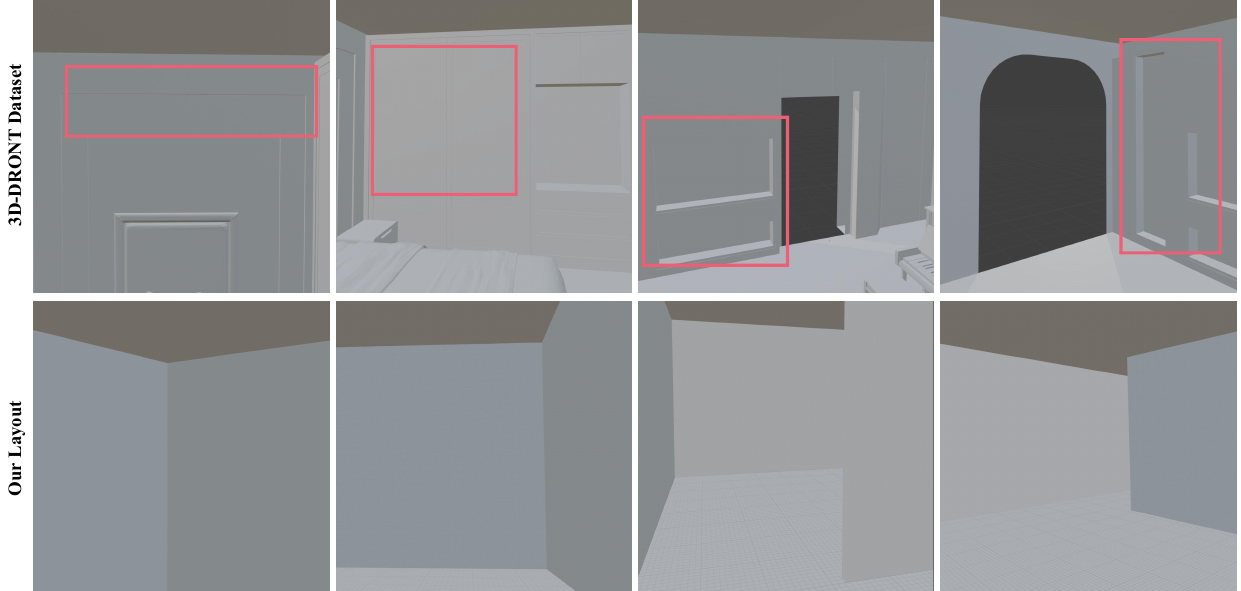}
    \caption{Comparisons of 3D layout mesh and scene mesh in 3D FRONT dataset. Our layout consists of simple planar structures. In contrast, intricate geometries on walls within 3D FRONT dataset.}
    \label{fig:layout_mesh}
\end{figure}

\paragraph{Multi-resolution Texture Field.}
Indoor scene texture generation methods can be categorized into two types: Inpaint-based methods\cite{roompainter, roomtex} and Optimization-based methods\cite{scenetex}. Inpaint-based methods directly inpaint texture map frame by frame conditioned on depth map. However, the inpainting results are strictly constrained by the depth map, and most environments(3D FRONT) utilized by existing methods\cite{roompainter, roomtex} contain complex geometric structures. As shown in Figure \ref{fig:layout_mesh}, due to the extreme flatness of our layout, inpaint-based methods applied to our scenes can only generate appearances with monotonous colors and missing textures. Optimization-based methods can address the aforementioned issues; however, due to gradient conflicts across different viewpoints, they often suffer from multi-view inconsistency.

Inspired by SceneTex\cite{scenetex}, we introduce multi-resolution texture field to encode appearance of environment. In particular, we encode texture featurs for all UV coordinate $q$ at each scale, can concatenate those features as the output UV embedding $\varepsilon(q)$ to represent all texture details. 

\paragraph{Normal-aware Cross-attention Decoder.}
Since the texture is optimized in image space, texture are often contrained by limited field of view and self-occlusion. As a result, the optimized texture often suffers from style inconsistency. SceneTex\cite{scenetex} decodes rgb value by cross-attending to all pre-sampled anchors in current instance. While SceneTex ensures style consistency, it overlooks a critical fact: regions with different normals often exhibit distinct texture characteristics. For example, the floor, walls, and ceiling of a room typically feature distinct styles. 

Therefore, we decode rgb value by cross-attending to the pre-sampled anchors which have the same normal. This improvement introduces minimal stylistic variation while ensuring that regions with different normals retain their distinct texture characteristics. Sec. \ref{ablation_experiments} conduct ablation study for normal-aware decoder.

\paragraph{Zigzag adaptive camera trajectory.}
\begin{figure}
    \centering
    \includegraphics[width=1.0\linewidth]{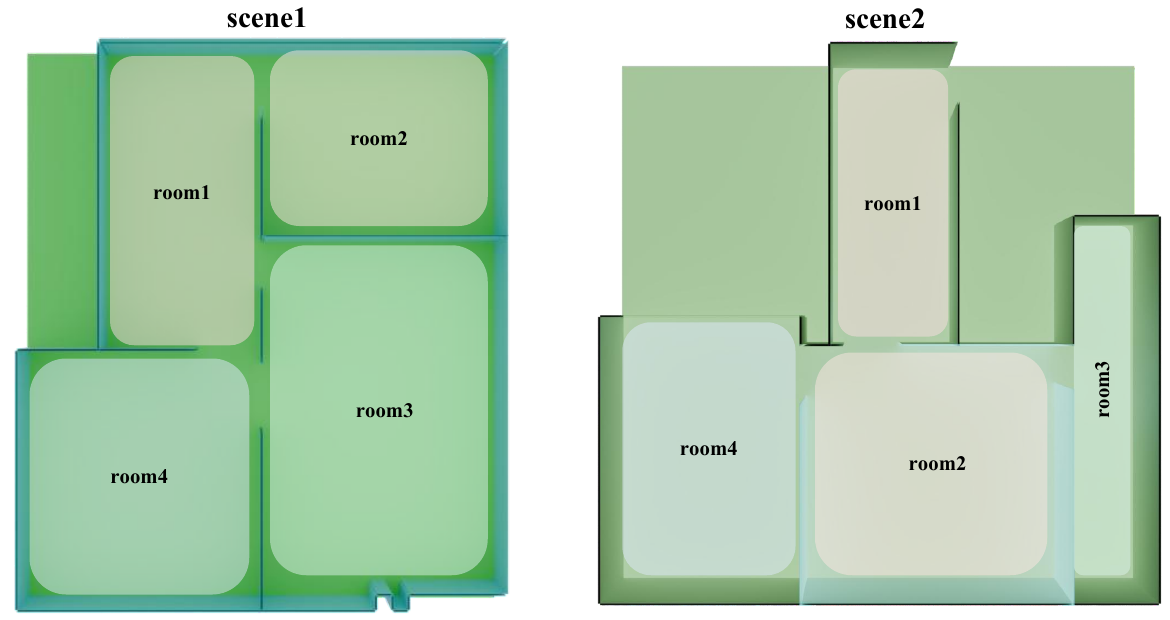}
    \caption{Room Scale. Our layout encompasses rooms of various scales, necessitating a camera trajectory that is sufficiently robust.}
    \label{fig:room_scale}
\end{figure}
As shown in Figure \ref{fig:room_scale}, Our layout includes rooms of different scales, requiring a robustly designed camera trajectory to cover all areas. Previous methods either require predefined camera trajectories\cite{SceneCraft, scenetex} or sphere camera trajectory(place the camera at the room's center for rotation)\cite{Dreamscene, roompainter, roomtex}. However, manually defining camera trajectories is inapplicable to scenes of all scales and requires substantial manual effort; moreover, sphere camera trajectory fails to accommodate all room with different scale as shown in Figure \ref{fig:camera_issue}. 

\begin{figure}
    \centering
    \includegraphics[width=1.0\linewidth]{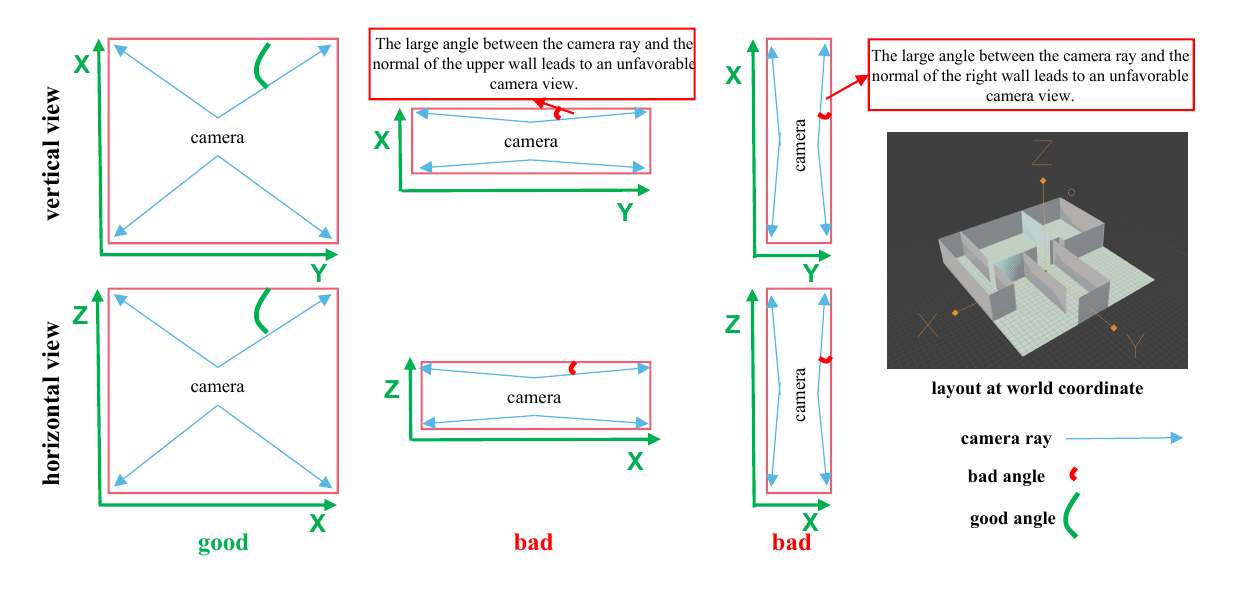}
    \caption{The issue of sphere camera trajectory. When the room has equal length and width, placing the camera at the center can capture a relatively good camera view (as shown on the left). However, when the room is relatively narrow (as shown on the right), the camera rays will deviate excessively from the environmental normals, resulting in an unfavorable viewing angle and lead to bad view.}
    \label{fig:camera_issue}
\end{figure}

\begin{figure}
    \centering
    \includegraphics[width=1.0\linewidth]{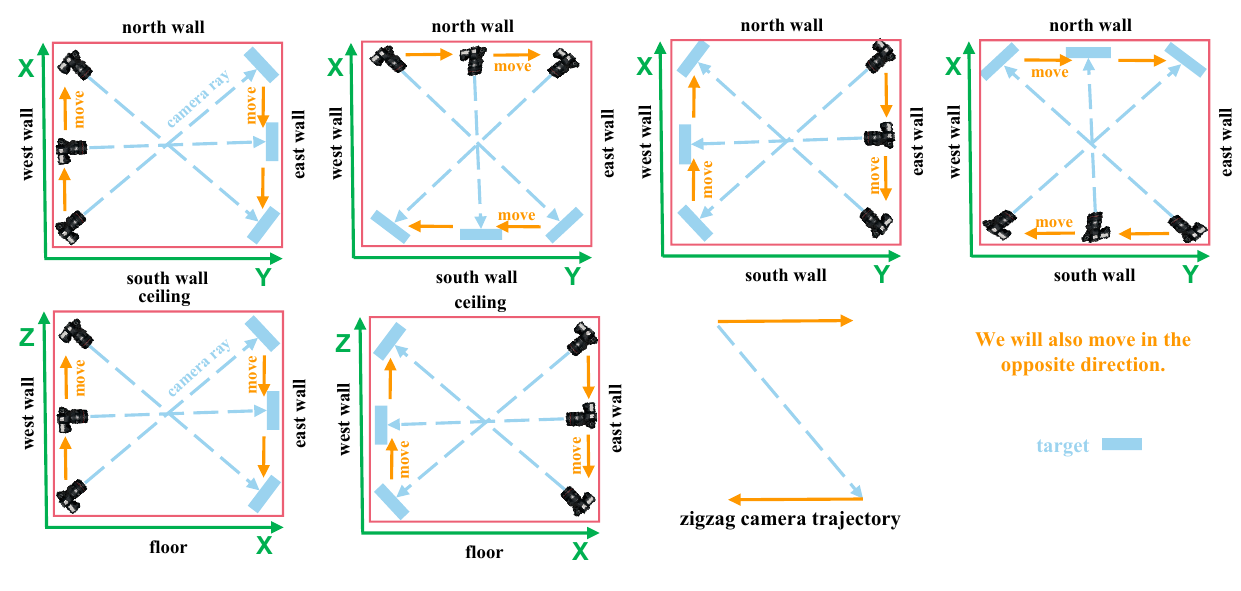}
    \caption{Zigzag camera trajectory. The first row is the camera trajectory viewed from the z-axis, the second row is the camera trajectory viewed from the y-axis. The camera moves along the wall, while the target moves in the opposite direction relative to the camera along the opposite wall. This approach offers three key advantages: (1) It prevents the short distance from causing diffusion models to fail in recognizing scene elements; (2) It ensures the angle between camera rays and target surface normals remains within acceptable limits; (3)It maximizes object coverage in the frame, avoiding texture monotony.}
    \label{fig:z_cam_trajectory}
\end{figure}

Therefore, we propose a zigzag adaptive camera trajectory that can accommodate indoor scenes of varying scales. As shown in Figure \ref{fig:z_cam_trajectory}, the camera’s xy coordinates move in the opposite direction to those of the target, while its height is inversely proportional to the target’s height(z axis). Three advantages of zigzag camera trajectory:
\begin{itemize}
    \item Maintaining a reasonable camera-target distance allows the diffusion model\cite{ddpm} to recognize scene semantics (e.g., floor, walls, ceiling), enabling style differentiation for semantic instances.
    \item Ensures the angle between camera rays and target surface normals remains within acceptable limits.
    \item Camera view can cover more furniture and environment elements, avoiding texture missing.
\end{itemize}
We conducted generation tasks on a total of 17 rooms across 4 houses, validating the effectiveness of our proposed method.

\paragraph{Optimization via CTS loss.}
We apply our CTS loss and zigzag camera trajectory for scene optimization. Training schedule:
\begin{itemize}
    \item warm up stage: we set the room center as the camera position. And randomly choose a furniture center as the target. Then we place all furniture into layout and render image. LCM\cite{lcm} used to optimize environment. 
    \item object remove stage: we select half of the furniture from the furniture list to incorporate into the layout for rendering, while simultaneously sampling camera positions from the 
    zigzag adaptive camera trajectory.
    \item noise removal stage: we remove the noise from CTS loss. 
\end{itemize}
Our method not only synthesizes photorealistic appearance but also preserves fine-grained surface textures. Furthermore, it autonomously generates sophisticated wall decorations including windows, curtains, and lamps. The noise removal is critical for environment optimization. And we conduct the ablation study in Sec. \ref{ablation_experiments}.

\paragraph{Several techniques}
We conduct ablation study in Sec. \ref{ablation_experiments}.

\subsubsection{Physical Editing}
We provide two way to perform physically plausible editing.
\begin{itemize}
    \item Thanks to the $L_{layout}$ constraint, our furniture is strictly confined within bounding boxes and tightly aligned with their edges. This enables direct utilization of the bounding boxes for physics-based simulation. We use the bounding box as proxy and = perform the physical simulation in blender\cite{blender} and export the translation and rotation matrix. 
    \item For more precise physics simulation, we employ RadeGS\cite{radegs} to extract the mesh, which is then utilized as a proxy for conducting the physical simulation.
\end{itemize}

\subsection{Discussion for Related Work} \label{Discussion for Related Work}
\subsubsection{Indoor Scene Generation}
There are some related works, we will explain the reasons for comparing or not comparing them.:
\begin{itemize}
    \item Generative Framework: only require a pre-trained diffusion model 
    `   \begin{itemize}
            \item Text2Nerf\cite{text2nerf} and Text2Room\cite{text2room}: both of them are inpaint methods. We compare one of them. 
            \item set-the-scene\cite{set_the_scene}: we follow the official guidelines and create the same layout for training. We compare this method.
            \item Controlroom3d\cite{controlroom3d}: controlroom3d generate indoor scene based on custom layout. And it did not open source. 
            \item DreamScene\cite{Dreamscene}: DreamScene introduces Formation Pattern Sampling (FPS) to balance semantic information and shape consistency, and a three-stage camera sampling strategy to improve the quality of scene generation. Dreamscene require predefined 3D layouts as input and only receives relative positional information of objects. We create the relative positional json file according to layout. We compare this method.
            \item SceneCraft\cite{SceneCraft}: SceneCraft trains a 2D diffusion model conditioned on bounding box image and employ a SDS loss for optimization. It also require pre-defined layouts and camera trajectory. we use nerfstudio to create the same layout for training. And we custom the camera trajectory. We compare this method.
        \end{itemize}
    \item Object Retrieval Framework: not only pre-trained diffusion model, but a large database of furniture. Therefore, if comparison is needed, we only compare the layout generation.
        \begin{itemize}
            \item Anyhome\cite{Anyhome}: the first object retrieval framwork that utilize LLM to generate multi-room scene. AnyHome is not fully open-source; we compare the layout generation with anyhome. 
                \begin{itemize}
                    \item AnyHome filters out some improperly placed objects, resulting in large blank areas.
                \end{itemize}
            \item InstructScene\cite{InstructScene}: InstructScene trains a diffusion model to generate layout for single room generation. We compare the layout generation with InstructScene.
                \begin{itemize}
                    \item Only tackle single room. 
                    \item Diversity in the generated layouts is reduced due to the limited size of the training dataset.
                \end{itemize}
            \item Architect\cite{Architect}: Architect first apply a diffusion model to inpaint an empty scene, then pre-trained depth estimation models to lift the generated 2D image to 3D space. We compare the large furniture layout generation.
                \begin{itemize}
                    \item The masked area is usually located in the center, causing the model to tend to generate objects in the center of the room, with almost no objects along the walls.
                    \item There is no strict constraint on the spatial relationships between bounding boxes, allowing for potential overlaps between different bounding boxes.
                \end{itemize}
            \item Holodeck \cite{holodeck}: holodeck  leverages a large language model for common sense knowledge about what the scene might look like and uses a large collection of 3D assets from Objaverse to populate the scene with diverse objects. 
                \begin{itemize}
                    \item Holodeck tends to position objects along walls, resulting in less varied layouts.
                \end{itemize}
        \end{itemize}
\end{itemize}

\subsubsection{Object Generation}
\begin{itemize}
    \item Consistent3D\cite{consistent3d}: Consistent3d explores the ODE deterministic sampling prior for 3D generation. However, Consistent3d leverage nerf\cite{nerf} as 3D representation that ray tracing methods can not incorporate into our framework. Additionally, Consistent3d is a two stage method and our method is one step. Hence, we did not compare this methods.
    \item CFD\cite{CFD}: CFD and Consistent3D initialize 3D model by MVDream\cite{mvdream}, and Our model initializes by point-e\cite{pointe}. we did not compare this methods.
    \item CCD\cite{CCD}: CCD shares the same insight as the Consistent3d\cite{consistent3d} and also incorporate consistency model\cite{cm}. We compare this methods.
    \item Luciddreamer\cite{Luciddreamer}: Variants of the SDS loss.
    \item DreamLCM\cite{DreamLCM}: A sds loss that use the latent diffusion model as based model. However, it did not open source.
    \item Vividdreamer\cite{vividdreamer}: A sds loss that use the latent diffusion model as based model. We compare this methods.
    \item DreamScene\cite{Dreamscene}: We also compare with it.
\end{itemize}

\subsubsection{Environment Optimization}
Although some works\cite{scenetex, roompainter, roomtex} address the generation of both environmental and furniture textures, our approach is limited to layout textures for planar surfaces only. Consequently, we conduct ablation studies to validate the effectiveness of our approach.

\begin{figure}
    \centering
    \includegraphics[width=1.0\linewidth]{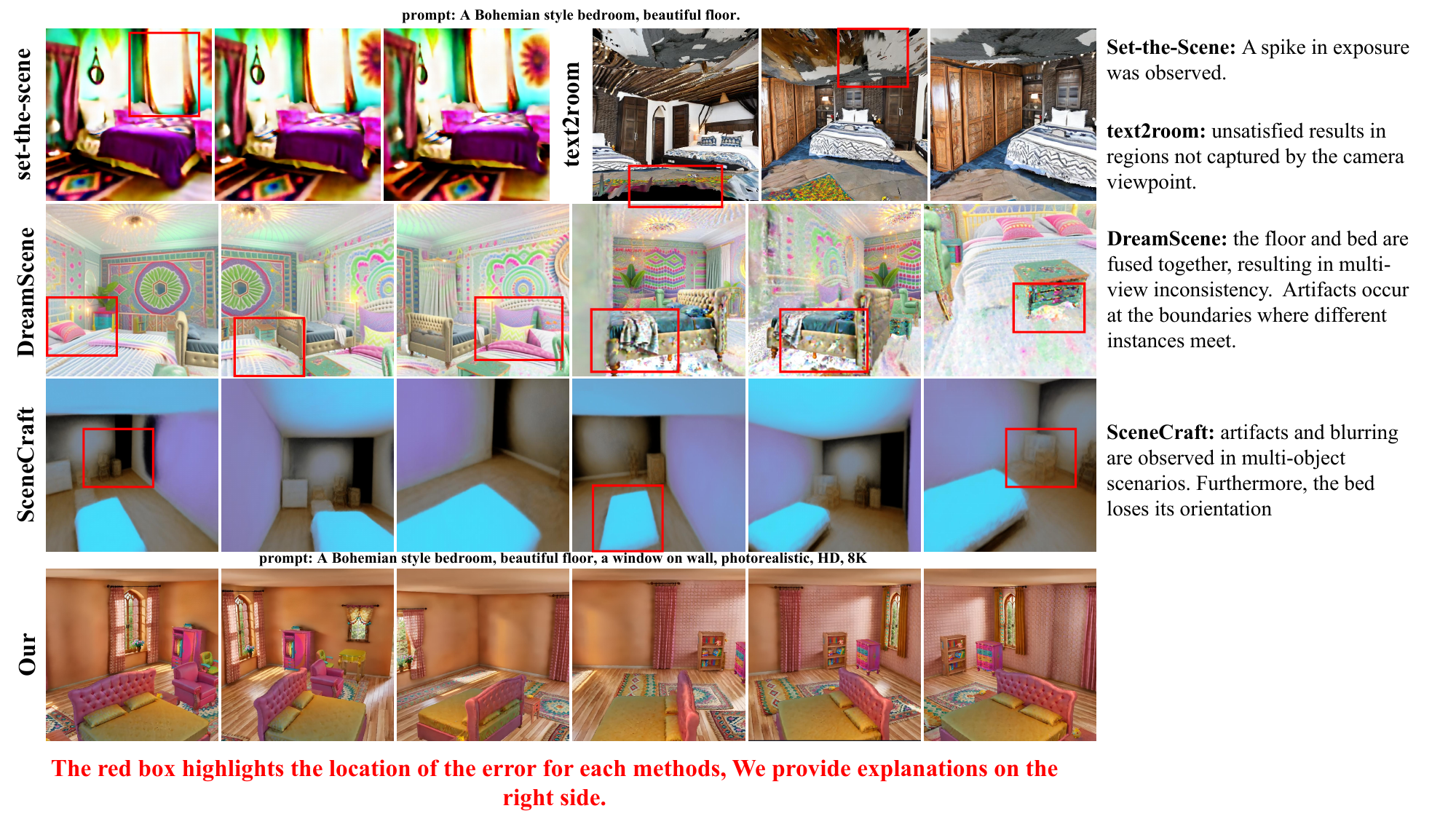}
    \caption{Comparisons of SceneLCM and baselines. }
    \label{fig:comparison_explain_1}
\end{figure}

\begin{figure}
    \centering
    \includegraphics[width=1.0\linewidth]{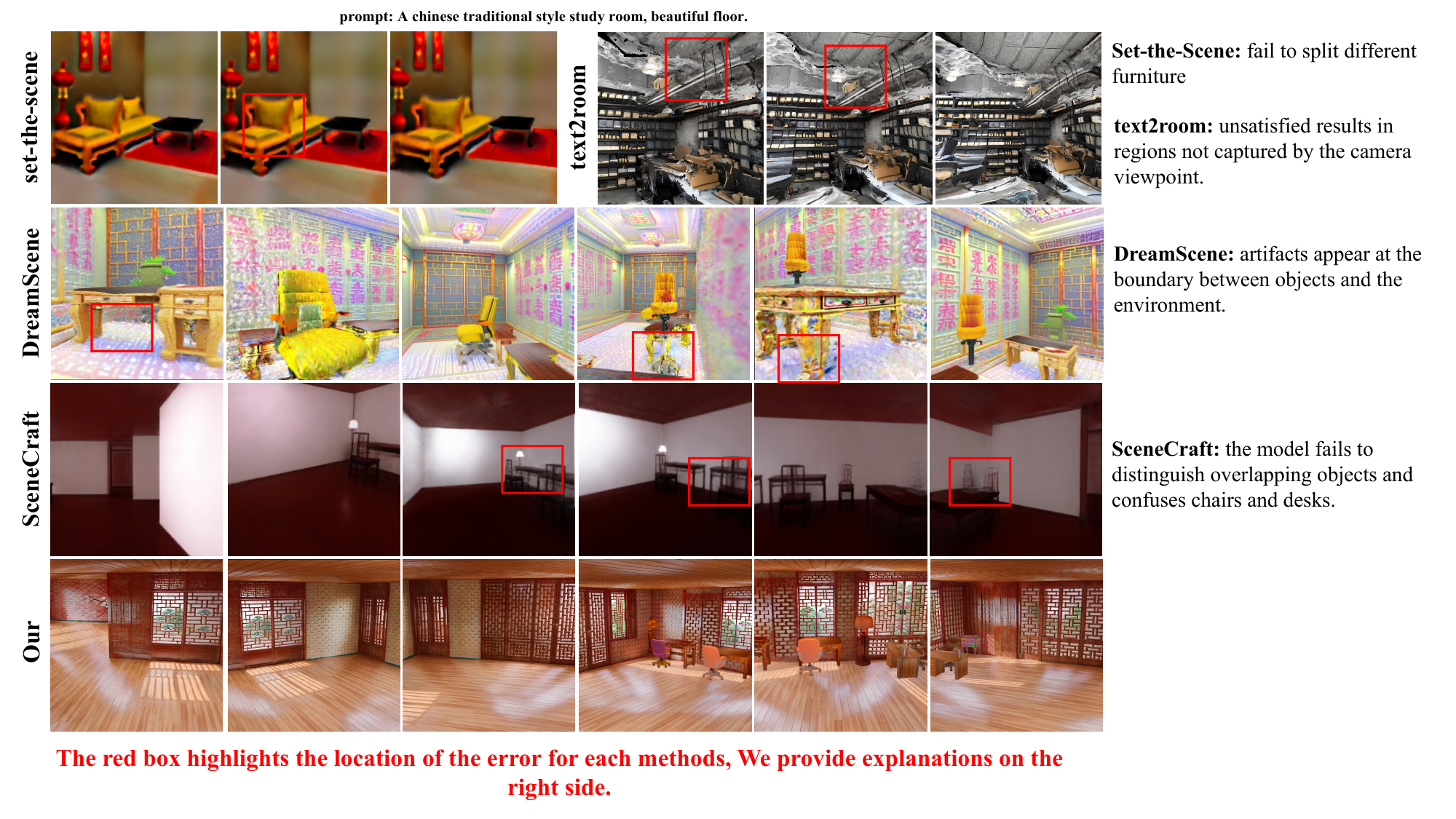}
    \caption{Comparisons of SceneLCM and baselines. }
    \label{fig:comparison_explain_2}
\end{figure}

\subsection{Explanation of Experiments} \label{Explanation of Experiments} 

We explain the comparison results in Figure 5 of regular paper. 

\begin{figure}
    \centering
    \includegraphics[width=1.0\linewidth]{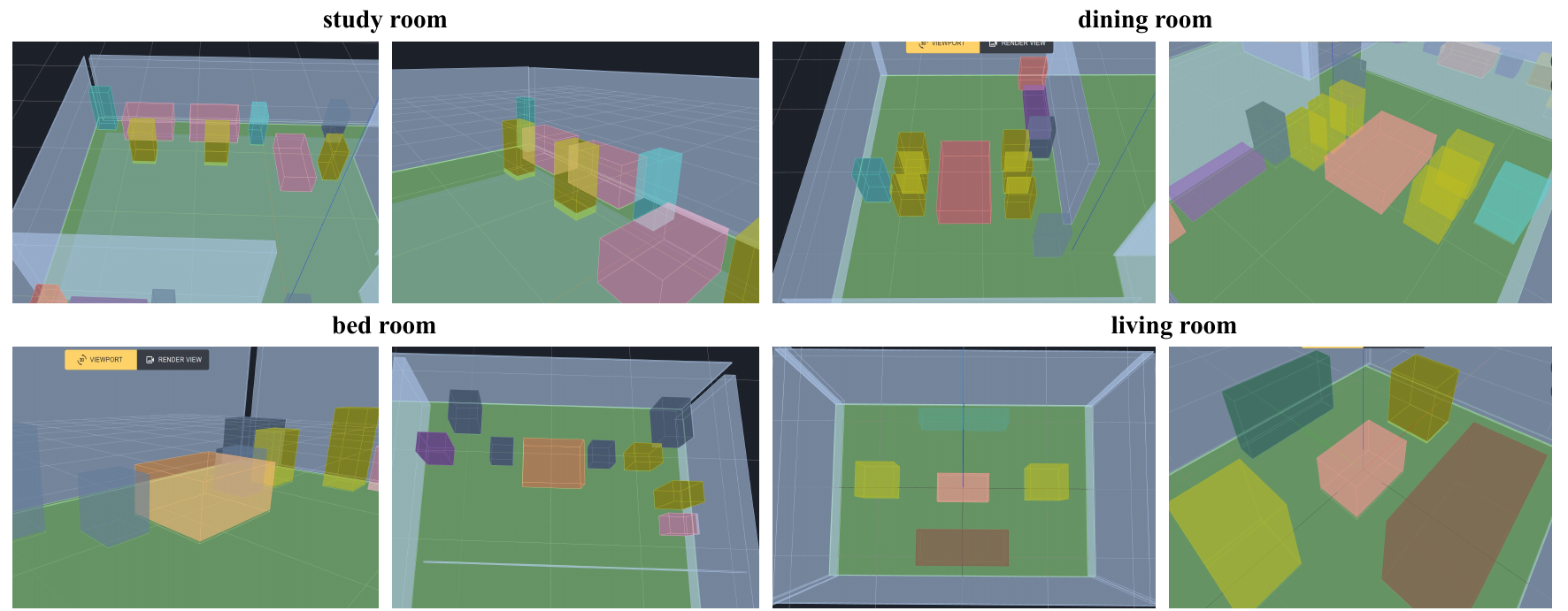}
    \caption{Layout of SceneCraft. }
    \label{fig:scenecraft_layout}
\end{figure}

As shown in Figure \ref{fig:comparison_explain_1} and Figure \ref{fig:comparison_explain_2}, Set-the-scene is nerf-composition methods that struggles with generalization across objects of varying scales and suffers from overexposure issues; text2room\cite{text2room} fails to generate surfaces in regions outside the camera views; DreamScene\cite{Dreamscene} lacks geometric priors of the environment, causing the environment and furniture to merge during optimization, resulting in artifacts and multi-view inconsistencies; SceneCraft\cite{SceneCraft} fine-tunes a 2D diffusion model for scene generation tasks; however, its inherent inability to model multi-object synergies leads to significant generation deficiencies in complex layout scenarios, resulting in incomplete structured scene outputs.
\begin{itemize}
    \item DreamScene\cite{Dreamscene}: \textbf{DreamScene's floorplan is automatically generated, creating the smallest possible enclosure that contains all objects. This is why DreamScene's layout differs from ours.}
    \item SceneCraft\cite{SceneCraft}: Our layout's scale differs from SceneCraft's, making direct replication impossible. We manually adapt our layouts (via NeRFStudio) to match SceneCraft's format. As shown in Figure \ref{fig:scenecraft_layout}. 
    \item However, despite we create the compatible layout for SceneCraft, it still generates unsatisfied results. 
\end{itemize}

\begin{figure}
    \centering
    \includegraphics[width=1.0\linewidth]{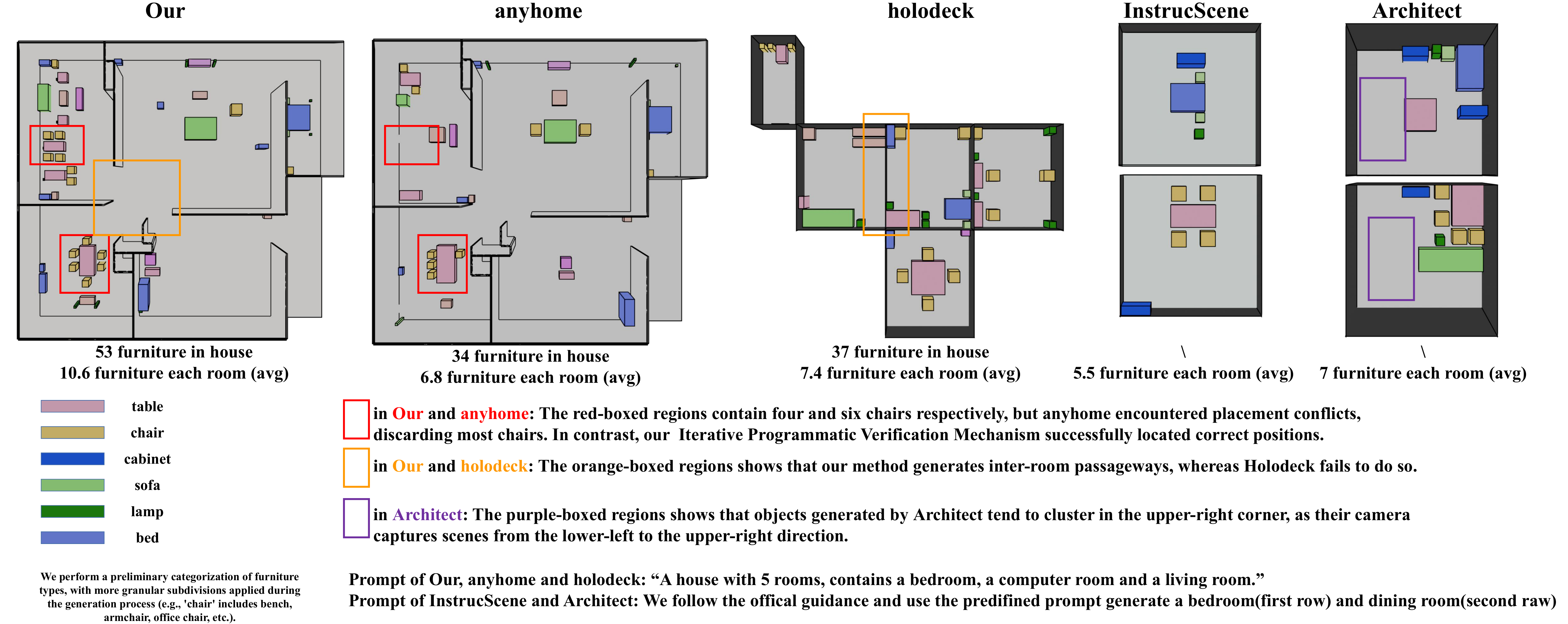}
    \caption{Layout Comparison with 4 basedline. AnyHome\cite{Anyhome} generates unreasonable layouts due to partial furniture omission. Holodeck\cite{holodeck} fails to create inter-room passageways and tends to align objects along walls. InstructScene\cite{InstructScene} and Architect\cite{Architect} are limited to single-room generation with simplistic layouts.
    }
    \label{fig:layout_comparison}
\end{figure}

\begin{figure}
    \centering
    \includegraphics[width=1.0\linewidth]{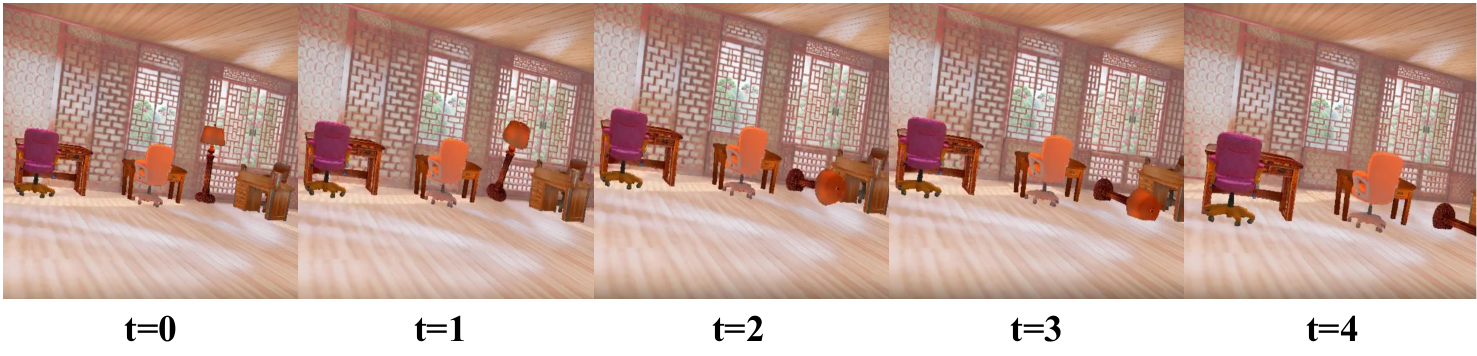}
    \caption{physical editing. Tilting the room by 30 degrees causes objects to slide downward due to gravity.}
    \label{fig:physical_editing}
\end{figure}

\begin{figure}
    \centering
    \includegraphics[width=1.0\linewidth]{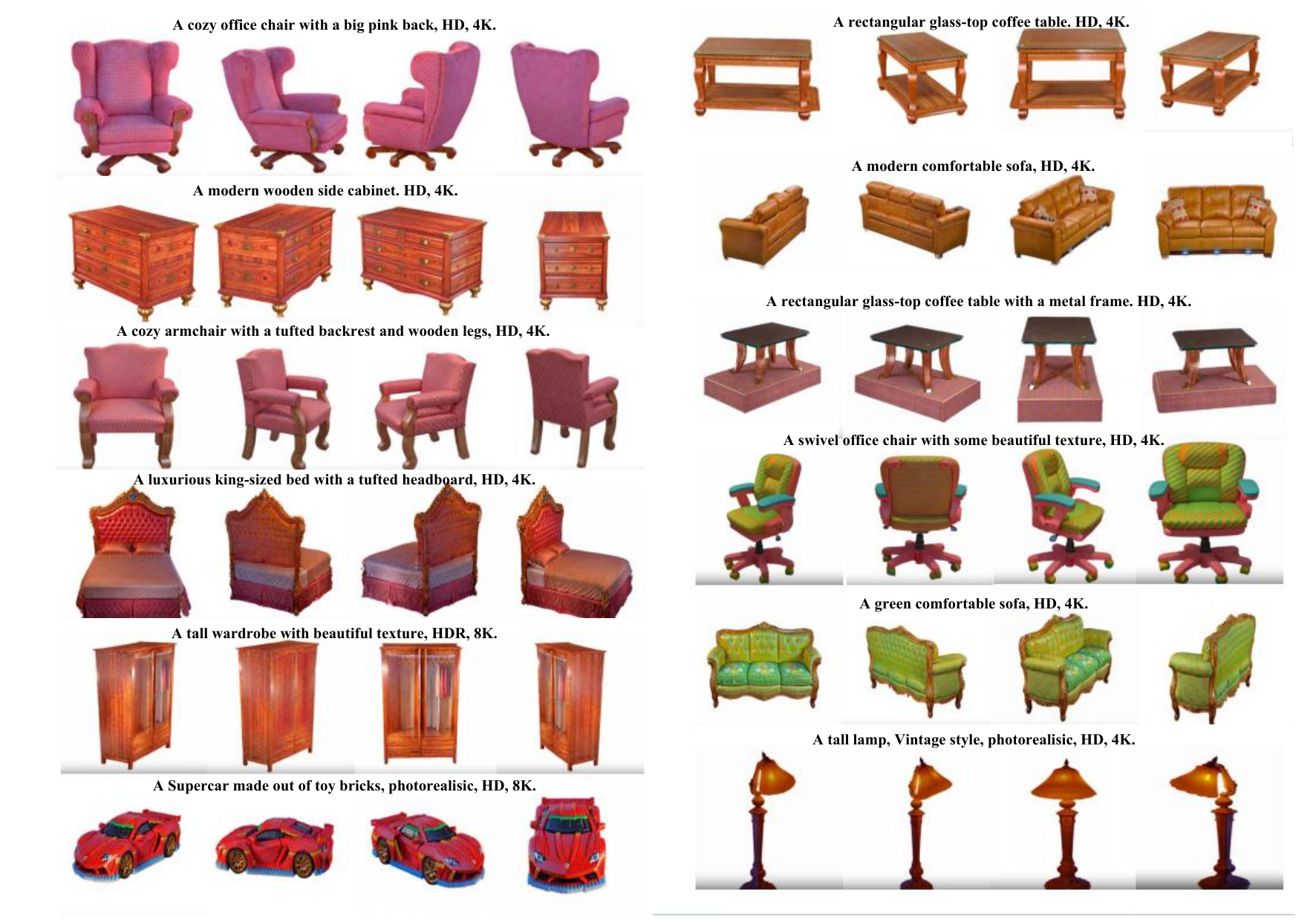}
    \caption{objects results}
    \label{fig:a1}
\end{figure}

\begin{figure}
    \centering
    \includegraphics[width=1.0\linewidth]{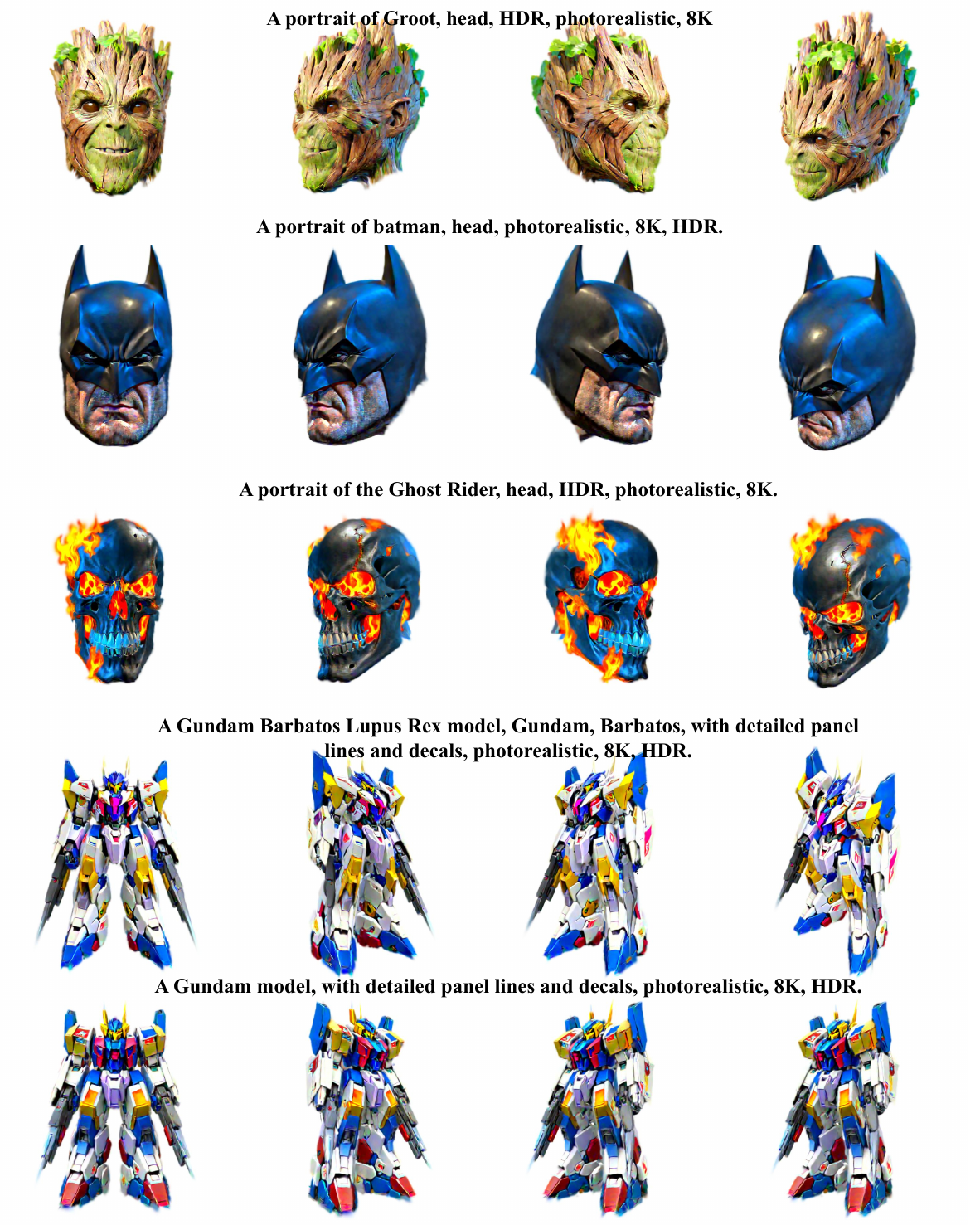}
    \caption{object resuluts}
    \label{fig:a2}
\end{figure}

\begin{figure}
    \centering
    \includegraphics[width=1.0\linewidth]{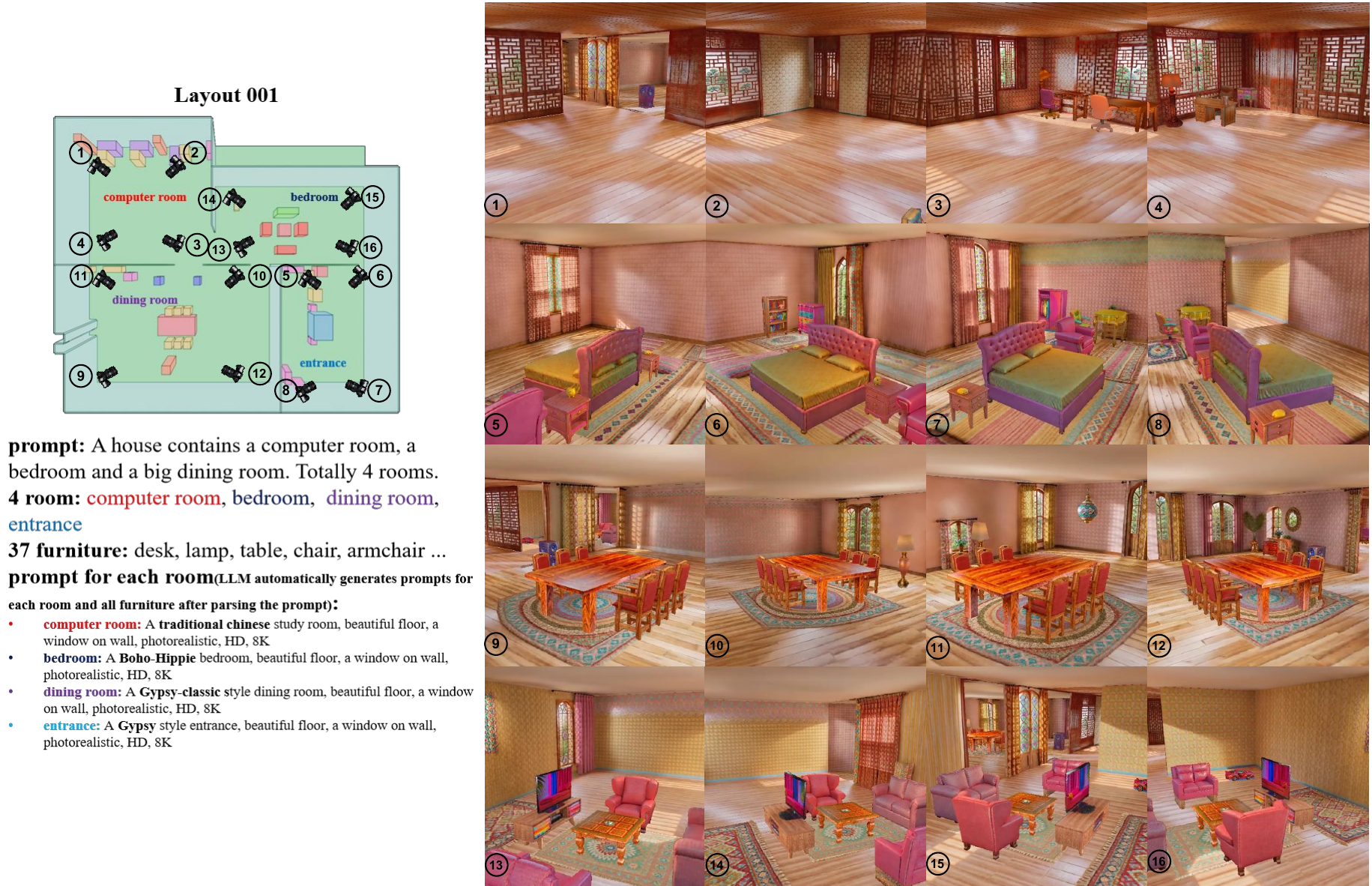}
    \caption{Additional Experiments Results for layout 1.}
    \label{fig:layout1}
\end{figure}

\begin{figure}
    \centering
    \includegraphics[width=1.0\linewidth]{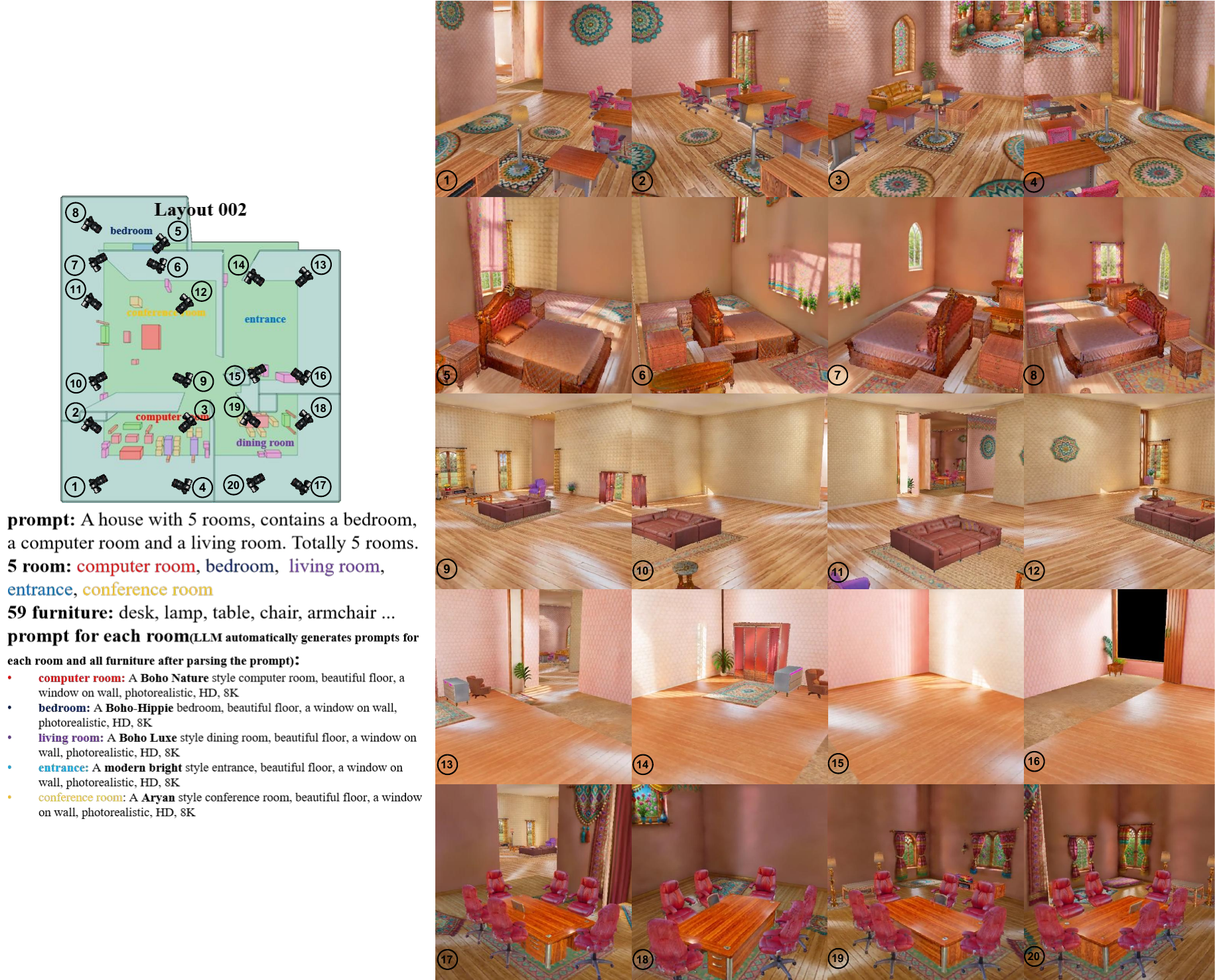}
    \caption{Additional Experiments Results for layout 2.}
    \label{fig:layout2}
\end{figure}

\begin{figure}
    \centering
    \includegraphics[width=1.0\linewidth]{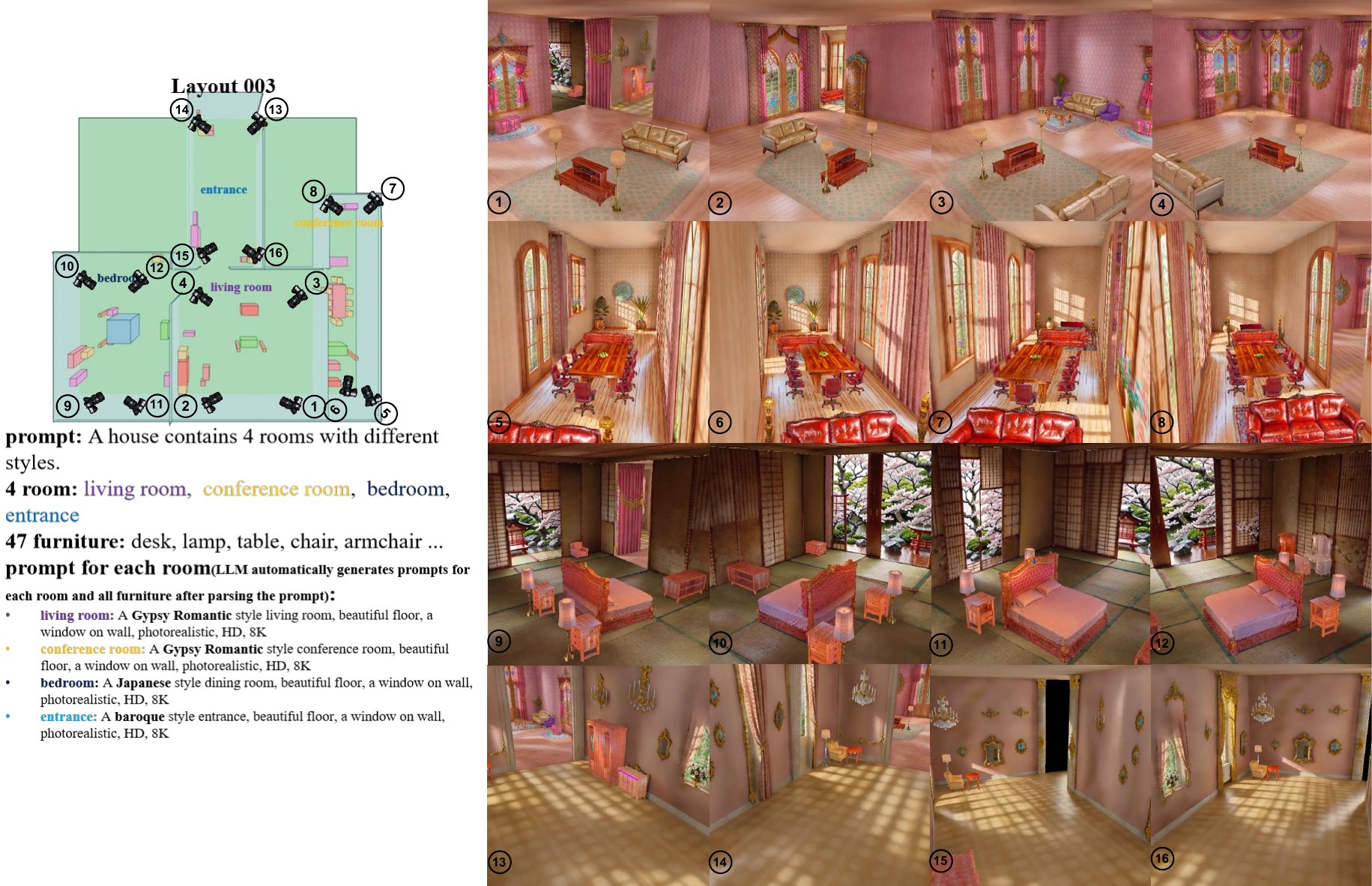}
    \caption{Additional Experiments Results for layout 3.}
    \label{fig:layout3}
\end{figure}

\begin{figure}
    \centering
    \includegraphics[width=1.0\linewidth]{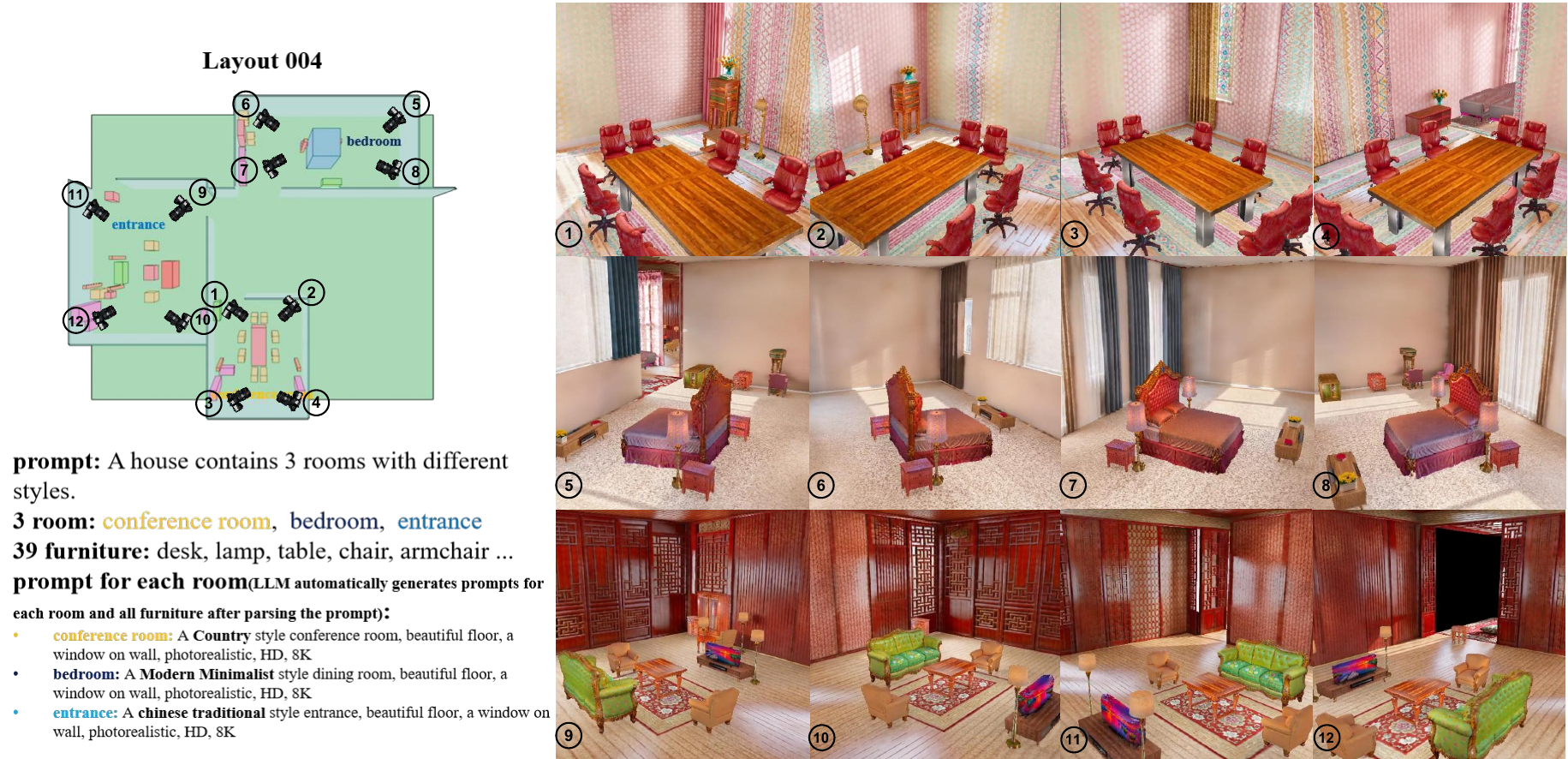}
    \caption{Additional Experiments Results for layout 4.}
    \label{fig:layout4}
\end{figure}

\subsection{Additional Experiments} \label{additional_experiments}
\begin{itemize}
    \item We conduct comparative experiments on layout generation.
    \item We conduct additional experiments on 2 house layout which contain 4 rooms and 3 rooms respectively. We conducted experiments across 4 houses, comprising a total of 16 rooms and over 182 furniture items. As shown in Figure \ref{fig:layout1},\ref{fig:layout2}, \ref{fig:layout3}, \ref{fig:layout4}.
    \item We provide additional visual results for both environments and objects.
    \item Physically plausible editing for multi-objects. As shown in Figure \ref{fig:physical_editing}, our model adheres to physical laws during the editing process.
    \item Editing
\end{itemize}

\subsubsection{Comparative Experiments on Layout}
To ensure fairness, all comparison models are using GPT-4. We use the same floorplan generative model as anyhome\cite{Anyhome}. Therefore, we first generate the floorplan, and then our method and AnyHome generate layouts on the same floorplan respectively. 
holodeck\cite{holodeck}, Architect\cite{Architect} and InstructScene\cite{InstructScene} generate a json file for saving bounding box of objects, we directly plot these bounding box in blender\cite{blender}. Additionally, InstructScene did not generate walls, which were manually created by us. 

As shown in Figure \ref{fig:layout_comparison}, our methods can generate more furniture and ensure the reasonable layout.

\subsubsection{More Results on House Generation}
We provide the video in our file. Due to time constraints, we cannot present the results here. The demonstration video has been included in the supplementary materials.

\subsubsection{Visual Results on Environment and Furniture}
We provide the video in our file. Furniture results as shown in Figure \ref{fig:a1}. We also provide some other results in Figure \ref{fig:a2}.

\begin{figure}
    \centering
    \includegraphics[width=1.0\linewidth]{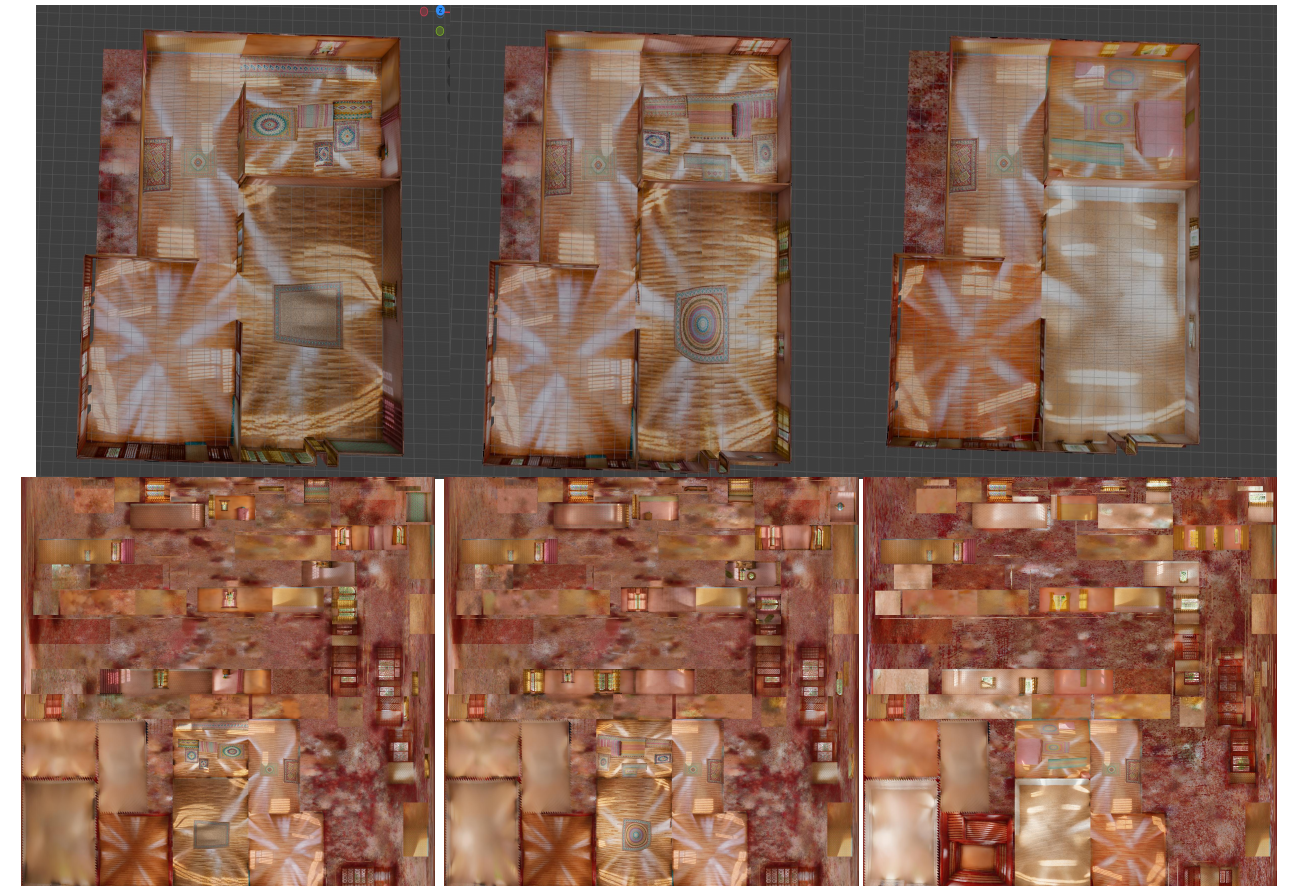}
    \caption{Environment Editing. We can export the texture map and editing the texture via textur e map.}
    \label{fig:env_editing}
\end{figure}

\subsubsection{Editing} 
\begin{itemize}
    \item Environment Texture Editing: As shown in Firgure \ref{fig:env_editing}.
    \item Furniture Editing:
        \begin{itemize}
            \item Create
            \item Delete
            \item Remove
            \item Update
        \end{itemize}
\end{itemize}

\subsection{Ablation Study} \label{ablation_experiments} 

\begin{figure}
    \centering
    \includegraphics[width=1.0\linewidth]{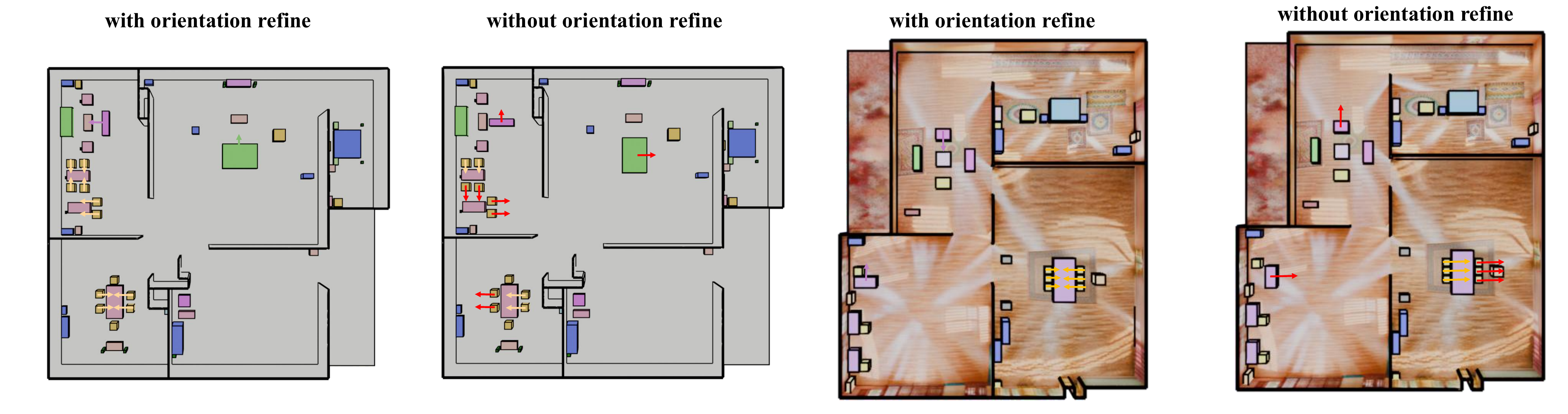}
    \caption{Ablation Study of Cluster-based Orientation Assignment. Red arrows indicate incorrect object orientations, as LLMs tend to generate uniform orientations for identical objects. Our method effectively corrects these inaccurate orientations.}
    \label{fig:ablation_orientation}
\end{figure}

\subsubsection{layout generation}
\begin{itemize}
    \item Iteractive Programmatic Verification Mechanism: As shown in Figure \ref{fig:layout_comparison}, without our proposed method, we would encounter the same issue as anyhome\cite{Anyhome}, where partial furniture is discarded, resulting in blank areas within the generated layouts.
    \item Cluster-based Orientation Assignment Strategy: As shown in Figure \ref{fig:ablation_orientation}, some inaccuracy orientation can be fixed after Orientation Assignment Strategy.
\end{itemize}

\begin{figure}
    \centering
    \includegraphics[width=1.0\linewidth]{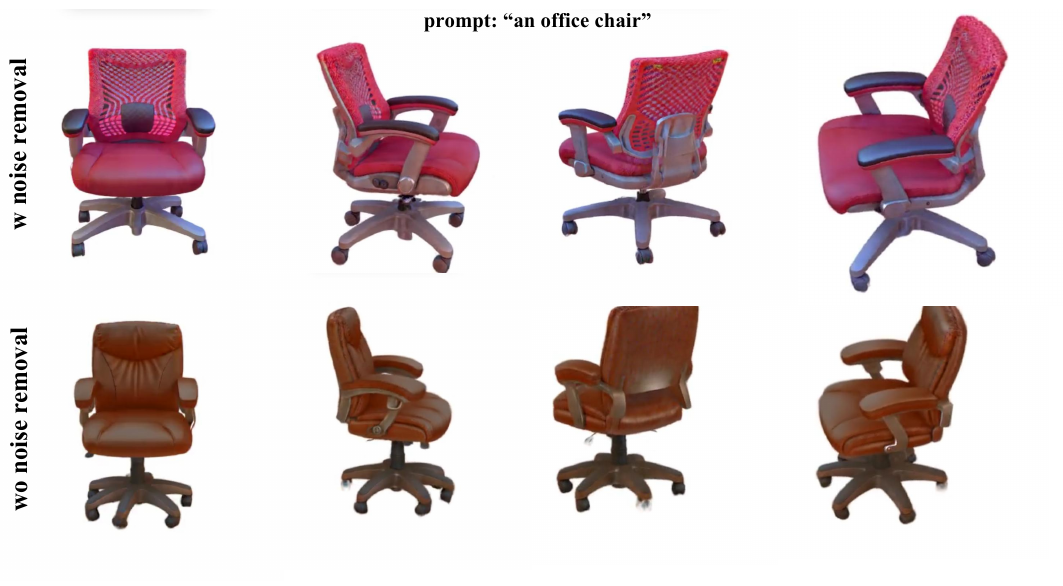}
    \caption{Ablation Study of noise removal. }
    \label{fig:ablation_removal_noise}
\end{figure}

\subsubsection{object generation}
We conduct an ablation study of noise removal, as shown in Figure \ref{fig:ablation_removal_noise}. Noise removal has texture with more detail and faster generation. To achieve comparable performance, adding noise requires 4,000 rounds (~36 minutes), while denoising necessitates 3,000 rounds (~28 minutes).

\begin{figure}
    \centering
    \includegraphics[width=1.0\linewidth]{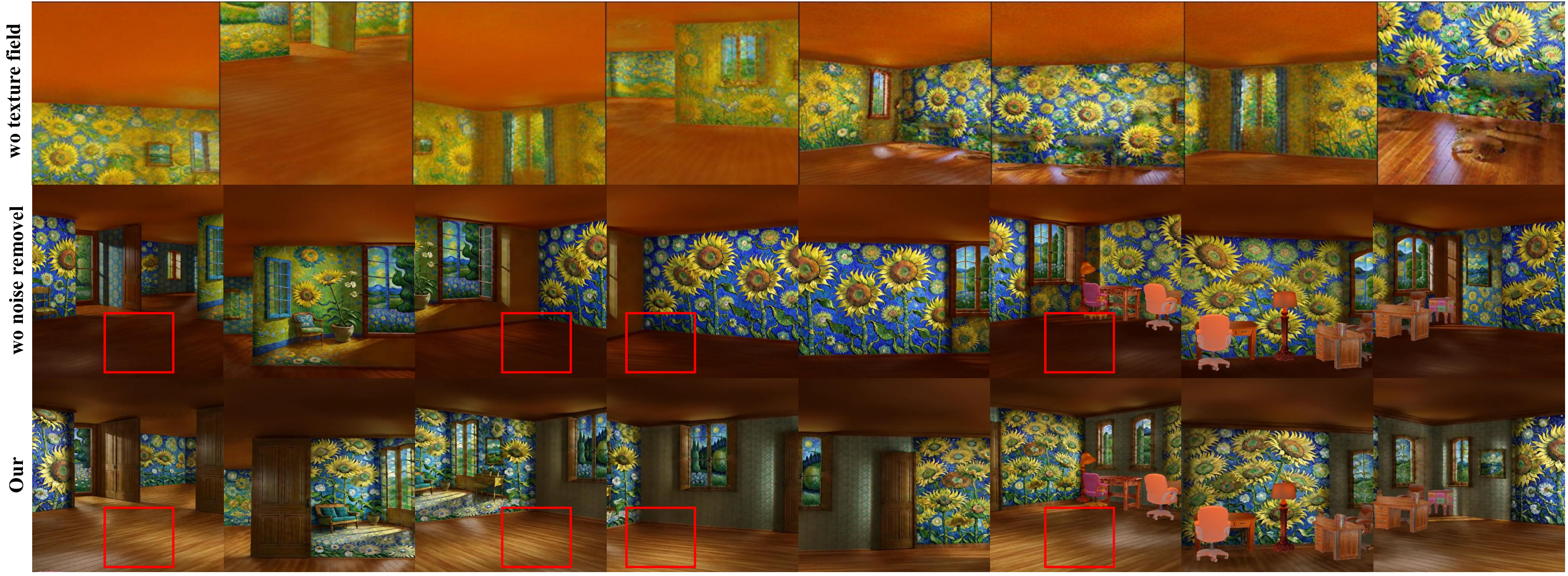}
    \caption{Ablation Study of noise removal and texture field on scene level(prompt: "A room in the style of Van Gogh"). Noise removal not only enhances the realism of generated scenes but also facilitates the generation of finer details, such as floor textures and patterns. Texture fields enable the generation of more intricate and detailed representations, while employing a unified neural network for UV coordinates significantly enhances spatial consistency. The first four images in Row 1 (left) and the subsequent four images (right) depict the same scene from different perspectives, revealing inconsistent floor styles.}
    \label{fig:ablation_texture_noise}
\end{figure}

\begin{figure}
    \centering
    \includegraphics[width=1.0\linewidth]{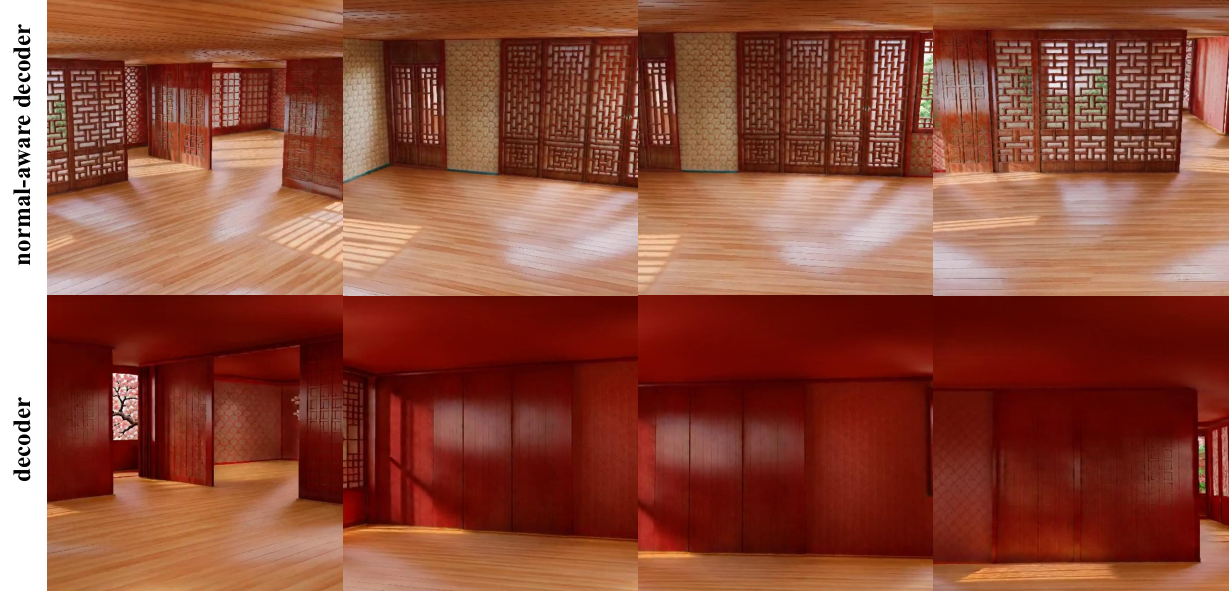}
    \caption{Ablation Study of normal decoder. Obviously, normal-aware decoder can generate much more details.}
    \label{fig:ablation_normal}
\end{figure}

\begin{figure}
    \centering
    \includegraphics[width=1.0\linewidth]{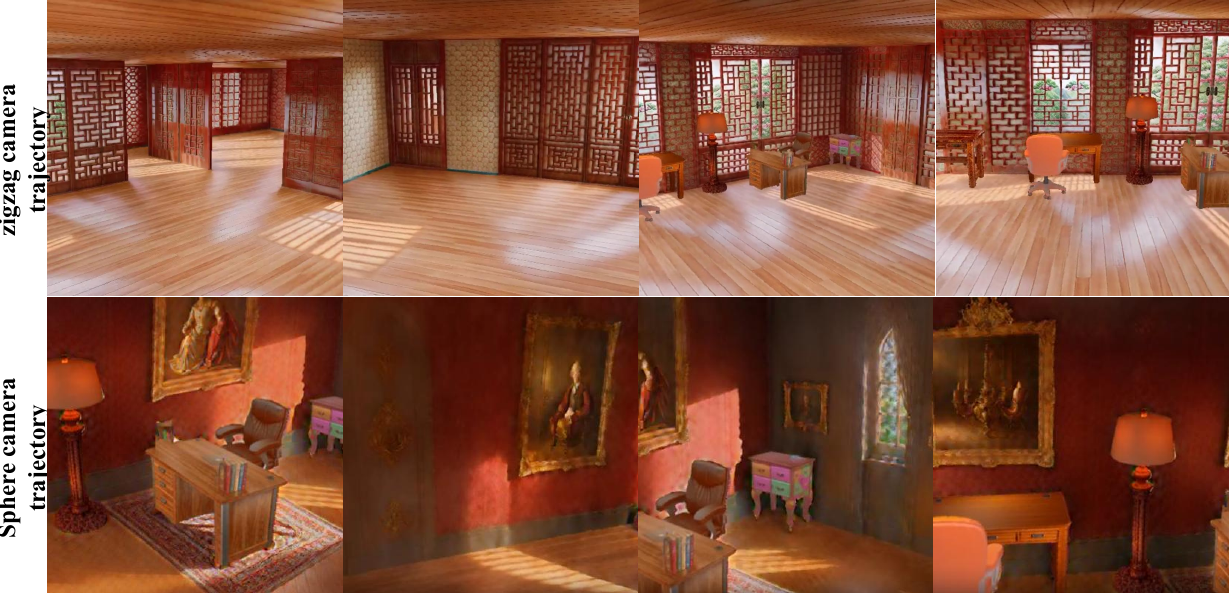}
    \caption{Ablation Study of camera trajectory. Placing the camera at the center results in the inability to capture combined entities (wall + floor + ceiling), preventing the model from recognizing the current entity's semantics and causing texture blurriness.}
    \label{fig:ablation_camera}
\end{figure}

\begin{figure}
    \centering
    \includegraphics[width=1.0\linewidth]{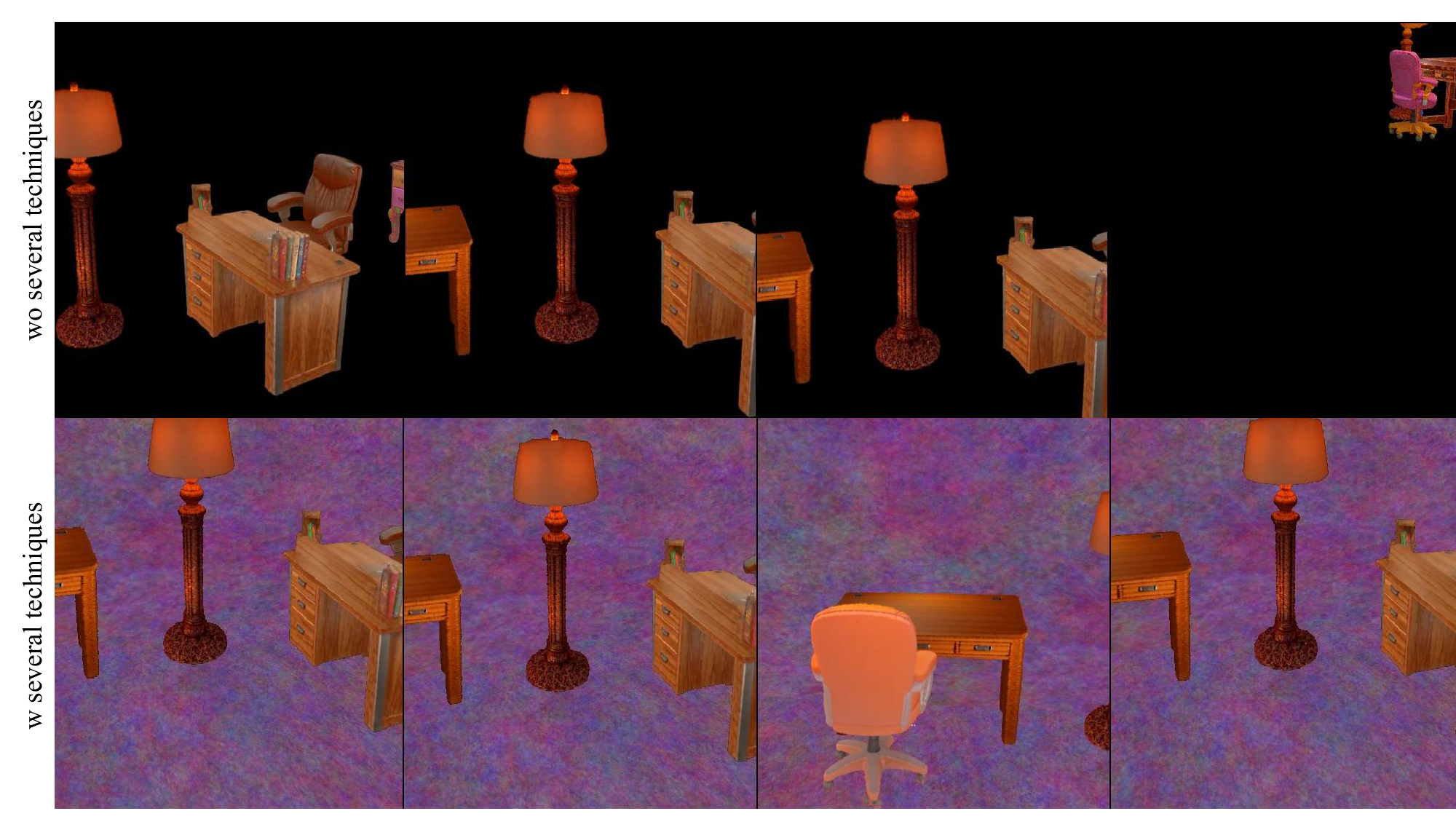}
    \caption{Ablation Study of servel techniques. These techniques introduce additional randomness into our initialization process, enabling successful optimization. Note: Final results are not presented here, as our model cannot converge without these techniques.}
    \label{fig:albaltion_stability}
\end{figure}

\subsubsection{environment optimization}
\begin{itemize}
    \item Multi-resolution texture field: As shown in Figure \ref{fig:ablation_texture_noise}. 
    \item Noise Removal: As shown in Figure \ref{fig:ablation_texture_noise}. 
    \item normal-aware decoder: As shown in Figiure \ref{fig:ablation_normal}, normal-aware decoder can generate much more details.
    \item zigzag camera trajectory: we use the sphere camera trajectory\cite{scenetex, Dreamscene} for comparison, as shown in Figure \ref{fig:ablation_camera}. Additionally, we conduct extensive experiments on rooms of diverse scales to validate the robustness of our method.
    \item several techniques: these techniques are prepared for numerical stability. As shown in Figure \ref{fig:albaltion_stability}.
\end{itemize}

\end{document}